\documentclass[letterpaper, 10 pt, conference]{ieeeconf}  

\IEEEoverridecommandlockouts                              

\usepackage{graphics} 
\usepackage{graphicx} 
\usepackage{mathptmx} 
\usepackage{times} 
\usepackage{amsmath} 
\usepackage{amssymb}  
\usepackage{color}
\usepackage{subfig}
\usepackage{epstopdf}
\usepackage{wrapfig}
\usepackage{adjustbox}
\usepackage{booktabs, makecell}
\usepackage{multirow}
\usepackage{times}
\usepackage{textcomp}
\usepackage{floatrow}
\DeclareMathAlphabet{\mathcal}{OMS}{cmsy}{m}{n}
\usepackage[normalem]{ulem}

\newcommand{\norm}[1]{\left\lVert#1\right\rVert}
   
\long\def\SUM{{\em SUM }}

\title{\LARGE \bf
SUM: Sequential Scene Understanding and Manipulation
}

\author{Zhiqiang Sui \hspace{0.5cm} Zheming Zhou \hspace{0.5cm} Zhen Zeng \hspace{0.5cm} Odest Chadwicke Jenkins
\thanks{Z. Sui, Z. Zhou, Z. Zeng and O.C. Jenkins are with the Department of Electrical Engineering and Computer Science, University of Michigan, Ann Arbor, MI, USA, 48109-2121
        {\tt\small [zsui|zhezhou|zengzhen|ocj]@umich.edu}}
}

\begin{document}

\maketitle
\thispagestyle{empty}
\pagestyle{empty}

\begin{abstract}
In order to perform autonomous sequential manipulation tasks, perception in cluttered scenes remains a critical challenge for robots. In this paper, we propose a probabilistic approach for robust sequential scene estimation and manipulation - Sequential Scene Understanding and Manipulation ({\em SUM}). \SUM considers uncertainty due to discriminative object detection and recognition in the generative estimation of the most likely object poses maintained over time to achieve a robust estimation of the scene under heavy occlusions and unstructured environment. 
Our method utilizes candidates from discriminative object detector and recognizer to guide the generative process of sampling scene hypothesis, and each scene hypotheses is evaluated against the observations. Also \SUM maintains beliefs of scene hypothesis over robot physical actions for better estimation and against noisy detections. We conduct extensive experiments to show that our approach is able to perform robust estimation and manipulation. 
\end{abstract}

\section{Introduction}

\begin{figure}[!t]
\includegraphics[width=0.98\columnwidth]{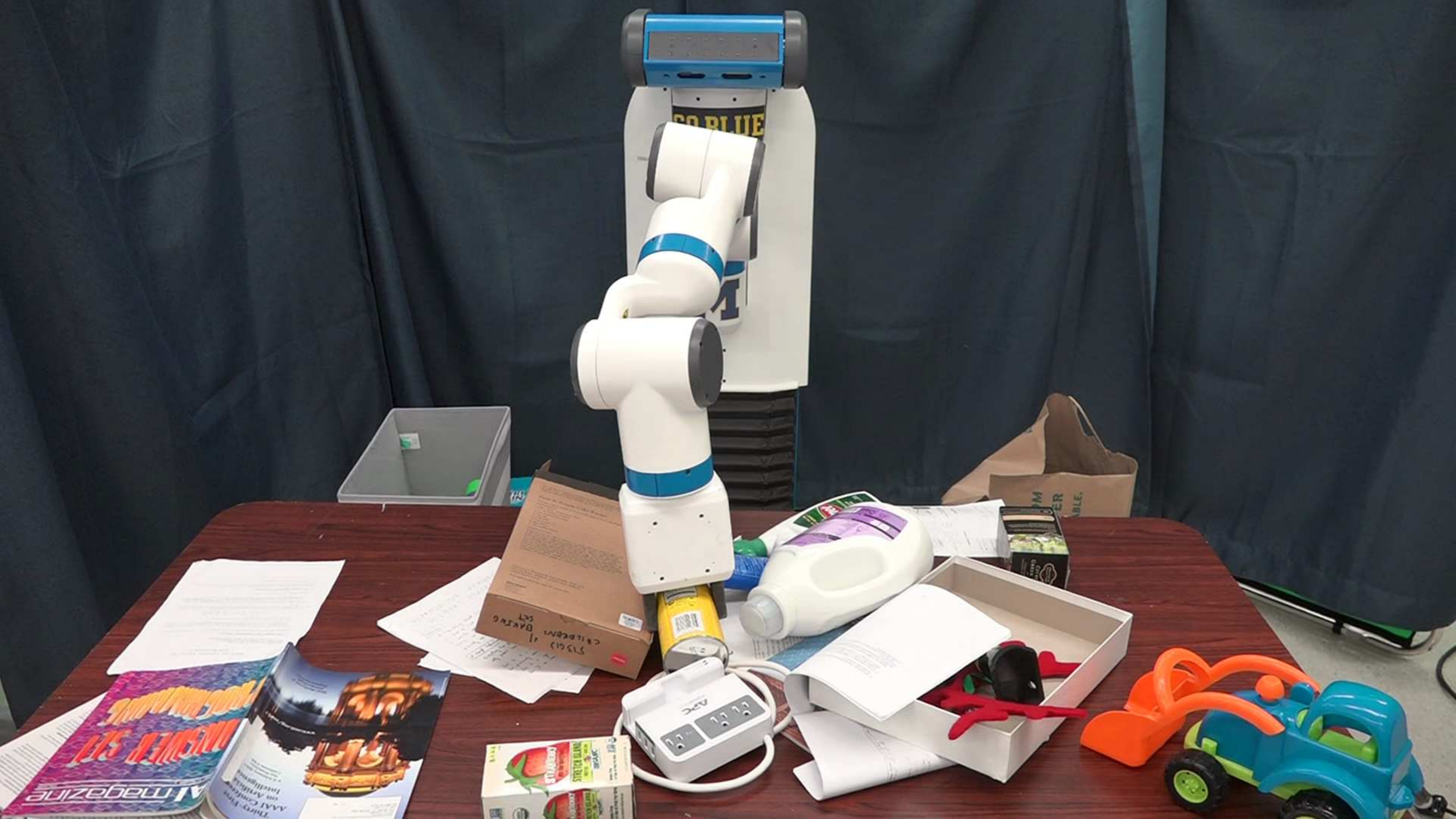}

\vspace{0.1cm}

\includegraphics[width=0.49\columnwidth]{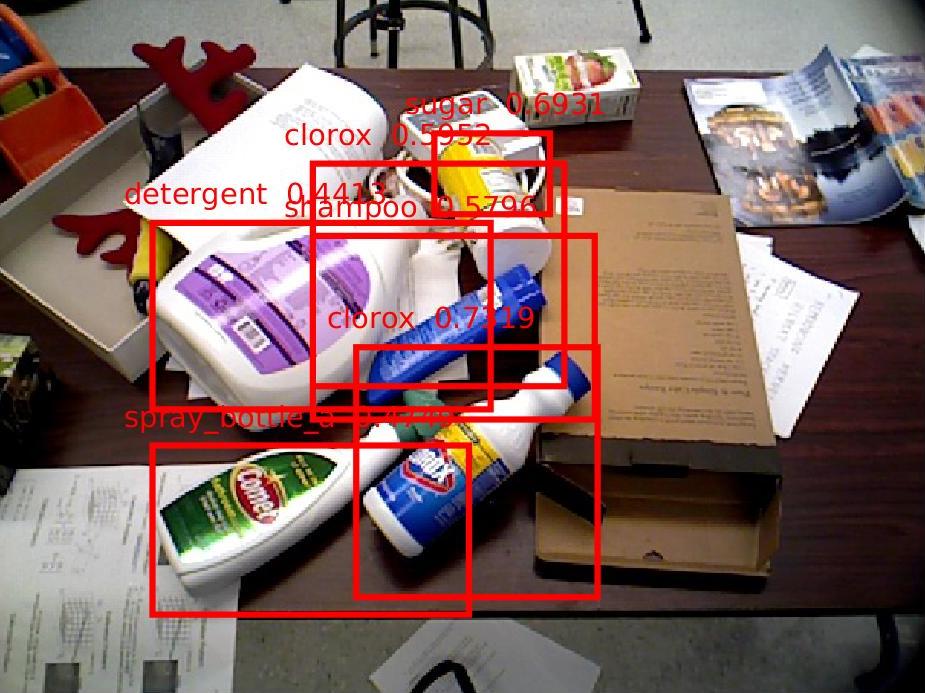}
\includegraphics[width=0.49\columnwidth]{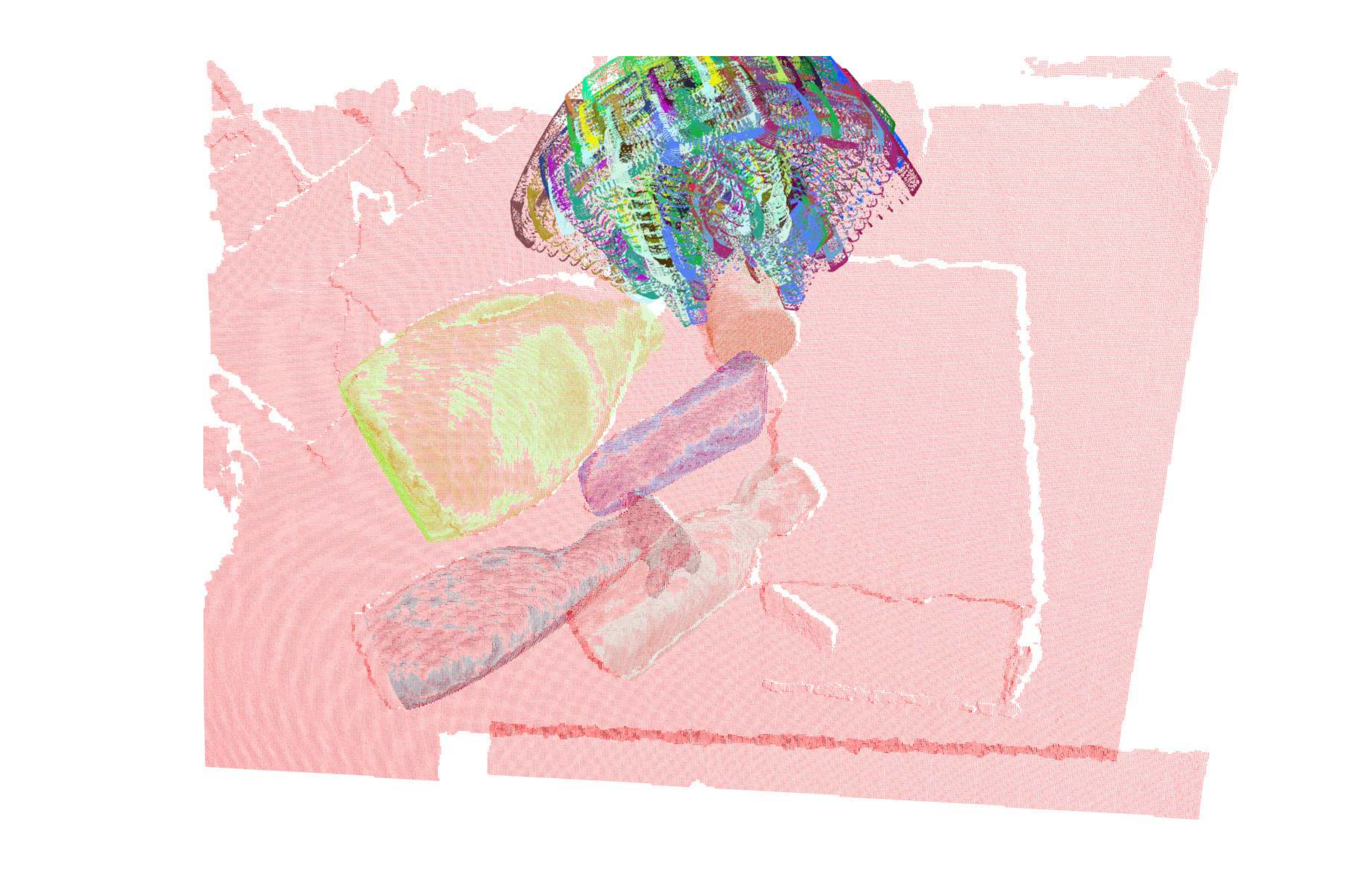}
\caption{\footnotesize 
(Top) a robot using \SUM for scene perception in the sorting of a cluttered set of objects on a table into cleaning (right bin) and non-cleaning categories (left bin).  \SUM performs perception by using (bottom left) RGB object recogition to inform (bottom right) sequential pose estimation from 3D point cloud observations and the feasible grasp poses on the manipulated object.
}
\label{fig:teaser1}
\end{figure}

Perception is a critical capability to enable purposeful goal-directed manipulation for autonomous robots, particularly in cluttered environments. 
Truly autonomous robot manipulators need to be able to perceive the world, reason over manipulation actions towards the goal, and carry out these actions in terms of physical motion.  
Closing this loop for autonomous scene-level manipulation is now within reach given the current advances in capable mobile manipulation platforms (such as the Fetch platform shown in Fig. \ref{fig:teaser1} and tractable task and motion planning.  Without the ability to perceive in common unstructured environments, however, autonomous manipulation will remain restricted to simulation and highly controlled environments. 

Previously, we addressed the problem of perception for goal-directed manipulation as {\em axiomatic scene estimation}~\cite{Suietal2015iros, Suietal2017ijrr}, sharing similar aims to existing work in scene estimation for manipulation of rigid objects~\cite{Narayanan-RSS-16,narayanan2016perch,Liuetal2015rcim,Johoetal2012,MOPED}.  These methods take a generative multi-hypothesis approach to robustly inferring a tree-structured scene graph, as object poses and directed inter-object relations, from cluttered scenes observed as 3D point clouds.  The inferred scene graph estimate can be expressed as parameterized axioms that allow for interoperable symbolic task and motion planning towards goals expressed as desired scenes in world space.  We posit this pattern of symbolic task-level reasoning using estimates from probabilistic perception will be as applicable to scenes for autonomous manipulation as it has been for autonomous navigation.

Existing formulations of axiomatic scene estimation impose several limiting assumptions that must be relaxed for viable autonomous manipulation.
First, existing axiomatic scene estimators assume the identification of objects observed in a scene are given or provided by an idealized object detection and recognition system. Object detection and recognition~\cite{jia2014caffe,RCNN,felzenszwalb2010object} has greatly improved towards feasible general use, due in part to the renaissance in convolutional neural networks.  However, such recognition methods remain subject to substantial and inherent errors in discriminating false positive and negative detections. As such, our robots will need to handle uncertainty due to such recognition errors in both its scene estimation and task execution.  

Second, scenes have been assumed to be static, where scene state at each moment in time is effectively decoupled sequentially from its past state. This assumption can be prohibitively costly in terms of computation, as the dimensionality of scene state space grows exponentially with the number of scene objects.  We posit that robots can manage this complexity by maintaining a belief of scene state over time informed by past beliefs, a manipulation process model, and current object detections, as well as incorporation of physical and contextual constraints~\cite{Desinghetal2016}. 

Lastly, existing scene estimation often assumes some scene structure, such as a canonical object orientation, a large flat surface support, or ``stacking'' as a single support surface per object. By maintaining belief sequentially and managing computational burden, our robots will be able to perform tractable inference in cluttered scenes with full six degree-of-freedom object poses and an arbitrary number inter-object contacts and supports.

In this paper, we propose Sequential Scene Understanding and Manipulation ({\em SUM}) as a model for scene perception from RGBD sensing in sequential manipulation tasks. 
\SUM considers uncertainty due to discriminative object detection and recognition in the generative estimation of the most likely object poses maintained over time.
Towards this end, \SUM utilizes the output of modern convolutional neural networks~\cite{RCNN} for object detection and recognition to guide the generative process of sampling scene hypotheses within a pose estimation process. Pose estimation is modeled as a recursive Bayesian filter~\cite{Dellaert_MCL, thrun2005probabilistic} to maintain a belief over object poses across a sequence of robot actions. We demonstrate the effectiveness and robustness of \SUM with respect to both perception and manipulation errors in a cluttered tabletop scenario for an object sorting task with a Fetch mobile manipulator.

\section{Related Work}
\subsection{Perception for Manipulation}
We summarize a relevant subset of existing work with respect to perception for manipulation. The PR2 interactive manipulation pipeline~\cite{ciocarlie2014towards} segments objects from a flat tabletop surface through clustering of surface normals. This pipeline can perform relatively robust pick-and-place manipulation for isolated, non-occluded, and non-touching objects. 
For cluttered scenes, Narayanaswamy et al.~\cite{narayanaswamy2011visual} perform pose and structure estimation of toy parts for flexible assembly of structures. MOPED~\cite{MOPED} is a framework for objects detection and pose estimation using clustering of features from multi-views. Joho et al.~\cite{Johoetal2012} use a probabilistic generative model to model the spatial arrangement of objects on a flat surface in the context of a table setting task. 

Narayanan et al. \cite{Narayanan-RSS-16,narayanan2016perch} are similar to our work on estimation where they integrate exhaustive global reasoning with discriminatively-trained algorithm to perform scene estimation. However, their work assume the identification of objects are given or provided by an idealized object detection and recognition system in order to perform A* search. 

In terms of sequential manipulation, the KnowRob system~\cite{Tenorth:2013} performs task-directed manipulation at the scale of entire buildings by integrating different knowledge sources from a perception system, an internal knowledge base and Internet repositories. Srivastava et al.~\cite{Srivastava:2013} perform jointly task and motion planning for goal-directed manipulation relying upon a hard-coded perception system. Cogsun et al.~\cite{Cosgun11IROS} performs sequential manipulation for placing objects in a scene where the objects are separating on a clear table. Similar to \SUM, Papasov et al.~\cite{papazov2012rigid} perform sequential scene estimation and manipulation through matching of known 3D geometries with an observed point cloud. However, this method takes a bottom-up approach using a RANSAC algorithm with Iterative Closest Point registration that neither requires nor uses a model of uncertainty.

\subsection{Object Recognition and Pose Estimation}

We consider two categories of traditional methods for model-based object recognition and pose estimation into two categories.  First, feature-based methods, also known as descriptor-based methods, aim to match key features in the models to the scenes. Key features can be comprised of local or global descriptors~\cite{aldoma2012point}.
Using local features~\cite{johnson1999using,rusu2008persistent,rusu2009fast}, the pose of the object is estimated by first matching a set of extracted features from object model to the scene.  Then, every matching pair will go through the filtering process to generate the final transformation.  In contrast, methods using global features~\cite{aldoma2012our,marton2011combined, rusu2010fast} attempt to match features with high resistance to the variance of the object pose. The object pose can be estimated by comparing those pose-preserving features from the training phase with the features computed from test scene.  A limitation of feature-based methods is that the estimation quality will degrade as the number of objects (and clutter) in the scene increases due to occlusion of key features.

Alternatively, generative methods (or {\it analysis by synthesis}) attempt to find the state estimate that best describes the observed sensor input through iterating over comparisons with state hypotheses rendered into sensor readings. The {\it Render-Match-Refine} paradigm of Stevens and Beveridge~\cite{stevens2000localized} applied iterative optimization method to find a rendering that best explain the input.  Their early work demonstrated Render-Match-Refine for 2D images, relying upon low-level feature extraction.  Recent work uses convolutional neural networks (CNNs) to compare rendered and observed images. Among this work, Krull et al.~\cite{krull2015learning} cast the CNN as a probabilistic model to output energy value, where as work by Gupta et al.~\cite{gupta2015inferring} directly output a coarse object pose and use ICP to refine it later. However, these methods do not address multi-object pose estimation such as in cluttered scenes. 

Similarly building on the renaissance in CNNs, object detection and recognition has greatly improved towards feasible general use.  The Region-based Convolutional Neural Network method (R-CNN)~\cite{RCNN} is a two-stage object detection system that integrates a high capacity CNN with bottom-up region proposal methods, which has demonstrated excellent detection accuracy.

\begin{figure*}
\includegraphics[width=0.9\linewidth]{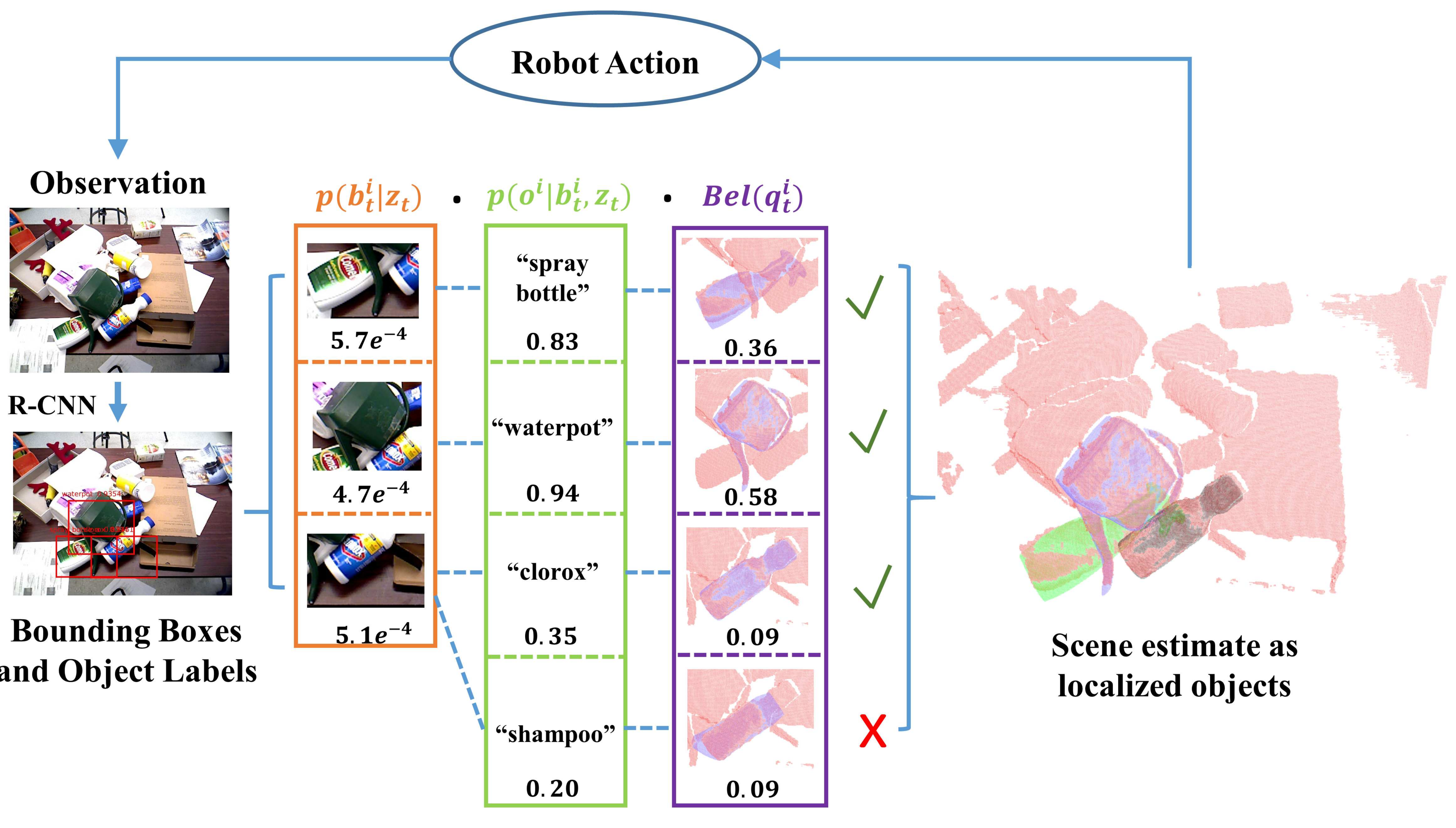}
\label{framework}
\caption{\footnotesize An overview of our \SUM framework. Given an observation of the scene, pre-trained R-CNN object detector and recognizer output bounding boxes and object labels along with their confidence. Assuming that every object is independent with each other, we estimate the state of the scene by estimating the state of each object individually. The numbers in the figure denotes the value of each term that composes to the posterior probability of a hypothesized object state. Multiple object pose estimator can originate from the same bounding box, for example, both the "shampoo" and "clorox" pose estimator originates from the same bounding box, and clorox is selected as the correct estimate since it has higher $p(x_t^i)$.  More detail is explained in Section \ref{sec:formulation} and Equation \ref{eq:conditional}.}
\end{figure*}

\section{Problem Formulation}\label{sec:formulation}

\begin{figure}
\centering
\includegraphics[width=0.8\columnwidth]{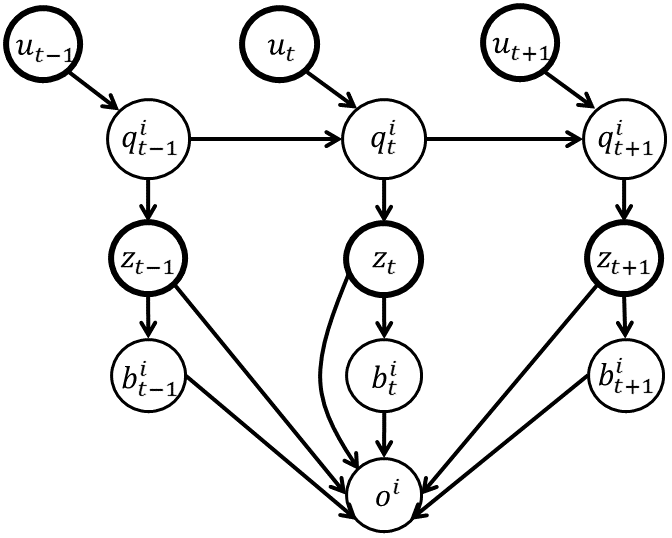}
\caption{\footnotesize Graphical model for estimating pose of a particular object $o^i$ given observations, actions. The bounding box and object label hypothesis at each frame is based on object detection, recognition and data association as explained in Section \ref{sec:methodology}.}
\label{fig:graphical_model}
\end{figure}

Given past RGBD observations $(z_0, \dots, z_t)$ and robot manipulation actions $(u_0, \dots, u_t)$, our aim is to estimate the scene as a collection of $k$ objects, with object labels $o$, 2D image-space bounding boxes $b_t$, and six DoF object poses $q_t$.  Note, $k$ is the number of objects {\it recognized} in the scene.  Object labels are strings containing a semantic identifier assumed to be a human-intuitive reference. In our experiments, the manipulation actions $u_t$ are pick-and-place actions, which will invoke a motion planning process. However, our formulation is general such that $u_t$ also applies to low level motor commands represented by joint torques, such as in scenarios for object tracking.
  The state of an individual object $i$ in the scene is represented as $x^i_t=\{q^i_t, b^i_t, o^i\}$. 
We assume that every object is independent of all other objects, which implies there will be only one object with a given label in the eventual inferred scene estimate.  Independence between objects allows us to state this scene estimation problem as:

\begin{equation}
p(x^1_t, \cdots, x^k_t|z_{0:t},u_{1:t}) = \prod_{i=1}^k p(x^i_t|z_{0:t},u_{1:t})
\end{equation}
where, for each object, the posterior probability is
\begin{align}
p(x^i_t|z_{0:t},u_{1:t})  \nonumber \\
=& p(q^i_t,b^i_t,o^i|z_{0:t},u_{1:t}) \nonumber \\
=& p(b^i_t|z_{0:t},u_{1:t})p(o^i|b^i_t,z_{0:t},u_{1:t})\cdotp \nonumber  \\
&  p(q^i_t|b^i_t,o^i,z_{0:t},u_{1:t})  \label{eq:chain_rule} \\ 
=& \underbrace{p(b^i_t|z_t)}_{detection}\underbrace{p(o^i|b^i_t,z_t)}_{recognition}\underbrace{p(q^i_t|b^i_t,o^i,z_{0:t},u_{1:t})}_{Bel(q_t^i)} \label{eq:conditional}
\end{align}

using the statistical chain rule and independence assumptions to yield Equations~\ref{eq:chain_rule} and \ref{eq:conditional}, respectively.  Equation~\ref{eq:conditional} represents the factoring of the scene estimation problem into object detection, object recognition, and belief over object pose.  The object detection factor $p(b^i_t|z_t)$ denotes the probability of object $i$ being observed within the bounding box $b^i_t$ given observation $z_t$. The object recognition factor $p(o^i|b^i_t,z_t)$ denotes the probability of this object having label $o^i$ given the observation $z_t$ inside the bounding box $b^i_t$. These distributions are generated as the output of a pre-trained discriminative object detector and recognizer that evaluates all possibilities.  The implementation of these detectors and recognizers is as explained in Section~\ref{sec:methodology}.
The pose belief factor for a particular object $o^i$ is modeled over time by a recursive Bayesian filter, as illustrated in Figure \ref{fig:graphical_model}. The belief over the object pose $q^i_t$ at time $t$ is estimated as:

\begin{multline}\label{eq:bayes_filter}
\begin{split} 
&Bel(q^i_t) \propto \\
&\underbrace{p(z_t|q^i_t,b^i_t,o^i)}_{observation \ model}\int_{q^i_{t-1}}\underbrace{p(q^i_t|q^i_{t-1},u_t,b^i_t,o^i)}_{action \ model}Bel(q^i_{t-1})dq^i_{t-1}
\end{split}
\end{multline}

Further explanations of the observation likelihood function and the action model are in section \ref{sec:methodology}.

\subsection{Data Association}
Across all objects, this Bayesian filtering framework also requires a data association process to correspond previous object estimators with the current detection and recognition.  Data association for \SUM maintains independent filters for each possible object, which are spawned or terminated based on object detection and recognition.  At the initial instance of time $t=0$, the number of objects $k$ is estimated by thresholding on the detection and recognition results for the initial observation $z_0$.
 For each recognized object $o^i$ along with its bounding box $b^i_0$, we will assign an object pose estimator to localize the object within the region defined by $b^i_0$. When the robot manipulates the objects in the scene for the next action $u_1$, the objects poses change as a result. To decide within which region that each object pose estimator should continue to localize the object $o^i$, there is a data association stage where it is associated to a bounding box $b^i_1$ detected at time $t=1$ after the robot action. Thus after every robot action $u_t$, the robot receives a new observation $z_t$, a data association stage takes care of associating each object estimator $T^i$ with a bounding box $b_t$ detected in $z_t$. More details on data association and when to terminate or add an object estimator is discussed in section \ref{sec:methodology}.

\section{Methodology}\label{sec:methodology}
\subsection{Object Detection and Recognition}

The \SUM model above is agnostic to the specific algorithms for objects detection and recognition as long as distributions of possible object bounding boxes $b$ and labels $o$ can be generated.  Specifically, each proposed detection of an object will have an bounding box in the image space with a probability of belonging to one of $N$ object labels in the training database. 
For each generated bounding box at time $t$, we filter out an object proposal $o_l$, $1 \leq l \leq N$, if its confidence is smaller than a certain ratio $\sigma_{c}$ of the maximum confidence in this bounding box:
\begin{equation}
p(o_l |z_0, b) < \max_{o_l} p(o_l |z_0, b) \cdot \sigma_{c}
\end{equation}
where $\sigma_{c} \in (0 \dots 1)$.  
The number of recognized objects $k$ is determined by the above thresholding procedure.  An object pose estimator $T^i$ is associated to each recognized object $i$ and its corresponding bounding box $b^i_0$ and object label $o^i$ pair. 

\subsection{Particle Filtering}
Particle filtering is employed with each object estimator $T^i$ to infer the pose $q_t^i$ of object $i$.  A particle filter is a means of inference for the sequential Bayesian filter in Eq.~\ref{eq:bayes_filter} through an approximation consisting of $n$ weighted particles, $\{q^{i_j}_t, w^{(j)}_t\}^{n}_{j=1}$. Weight $w^{(j)}_t$ for particle $q^{i_j}_t$ is expressed as:
\begin{equation} \label{eq:particle_filter}
\begin{split}
&Bel(q^i_t) \propto \\
&p(z_t|q^i_t,b^i_t,o^i)\sum_{j}p(q^{i_j}_t|q^{i_j}_{t-1},u_t,b^i_t,o^i)Bel(q^i_{t-1})
\end{split}
\end{equation}

as described by Dellaert et al.~\cite{Dellaert_MCL}. The initial belief of object pose is uniform. At each time instance, the weight of each hypothesis is computed, normalized to one, and resampled based on importance into an updated set of $n$ particles:
\begin{equation}
q_{t}^{i} \sim \sum_{j} w^{(j)}_{t-1}p( q_{t}^{i_j} | q^{i_j}_{t-1}, u_t)
\end{equation}
 
Before each robot action, we apply iterated likelihood weighting~\cite{mckenna2007tracking} to estimate the distribution of the object pose given the bounding box and the object label. This serves as a {\em bootstrap filter}, where the state transition in action model is replaced by a zero-mean Gaussian noise. 

Our observation likelihood function measures how well a particle's rendered point cloud explains the observation point cloud.
The observation model of this particle filter uses the z-buffer of a 3D graphics engine to render each particle $q_{t}^{i_j}$ into a depth image for comparison with the observation.  This depth image, represented as $\hat{z}_{t}^{(j)}$, is backprojected into a point cloud $\hat{r}_{t}^{(j)}$ in the camera frame to simulate the camera model.
The observation likelihood for each particle hypothesis with respect to the point cloud $r_t$ associated with the observation $z_t$ is then expressed as: 
\begin{equation}
    p(z_{t}|q_{t}^{i_j}, b_{t}^{i}, o^{i}) = \frac{\sum_{a, b \in \hat{r}_{t}^{(j)}} \text{INLIERS}(r_t(a, b), \hat{r}_{t}^{(j)}(a, b))}{N_{z_t}}
\end{equation}
\noindent where $a$ and $b$ are 2D indices in the rendered point cloud $\hat{r}_{t}^{(j)}$, $N_{z_t}$ is the total number of points in the observation point cloud and 
\begin{equation}
\begin{split}
  	\text{INLIERS}(p, p^{\prime}) =  
  &\begin{cases}
    1, & \text{if $\norm{p - p^{\prime}}_{2} < \epsilon$}\\
    0, & \text{otherwise} \\
  \end{cases}
\end{split}
\end{equation}
Thus, if the Euclidean distance between an observed point and a rendered point are within a certain sensor resolution $\epsilon$, total number of inliers will increment by 1.

A robot manipulation action is represented by $u(j, \phi_{pick}, \phi_{place})$.  This pick-and-place action is parametrized by the target object index $j$, object pick and place pose $\phi_{pick}, \phi_{place}$. For a particular object $o^i$, we use Gaussian components to model how the object pose $q^i_t$ will change from $q^i_{t-1}$ after a robot action $u_t(j, \phi_{pick}, \phi_{place})$,
\begin{equation}
\begin{split}
  p(q^i_t|& q^i_{t-1},u_t(j, \phi_{pick}, \phi_{place}),b^i_t,o^i) \\ \propto &\begin{cases}
    w_1 \mathcal{N}(\phi_{place}, \sigma_1^2) + w_2 \mathcal{N}(q_{t-1}^i, \sigma_2^2), & \text{if $i=j$}\\
    \mathcal{N}(q_{t-1}^i, \sigma_3^2), & \text{if $i \neq j$}
  \end{cases}
\end{split}
\end{equation}
If the action $u_t$ is targeted on object $o^i$, then either the action succeeds and the object is moved to the place pose $\phi_{place}$ with uncertainty characterized by $\sigma_1^2$, or the action fails and the object stays at its previous pose $q_{t-1}$ with uncertainty characterized by $\sigma_2^2$. If the action $u_t$ is not targeted on object $o^i$, then we assume that the object stays at its previous pose $q_{t-1}^i$ with uncertainty characterized by $\sigma_3^2$.
In cases where the action fails due the manipulated object being accidentally mishandled, the new pose $q_t$ far from its previous pose $q_{t-1}$ or its expected pose from manipulation success. 
This possibility is currently not modeled.  Instead, data association will handle this object through the spawning of a new estimator. 

\subsection{Data Association}
Data association is needed to associate each currently detected object bounding box with at most one object estimator at each moment in time. 
We use a greedy algorithm for our data association problem, which yields similar results at lower computational cost compared to the Hungarian algorithm (as reported by Breitenstein et al.~\cite{breitenstein2009robust}). First, a matching score matrix $S$ of every pair $(T^i, b_t^l)$ of object estimator and bounding box is calculated, with the matching score defined as
\begin{equation}
  s(T^i, b_t^l) = IoU(b_{t-1}^i, b_t^l)p(b_t^l|z_t)p(o^i|b_t^l,z_t)
\end{equation}
which consists of three factors: the overlap between $b_t^l$ and $b_{t-1}^i$ by Intersection over Union (IoU), the likelihood of an object to be in bounding box $b_t^l$, the likelihood of object $o^i$ inside the bounding box $b_t^l$.  The pair $({T^i}^\ast, {b_t^l}^\ast)$ with the maximum score in $S$ is selected as an established association.  The rows and columns belonging to the object estimator ${T^i}^\ast$ and the bounding box ${b_t^l}^\ast$ are removed from $S$. This process is repeated until no further pairing is possible. In the end, we only keep the established associations with a matching score above a chosen threshold.
A new object estimator is spawned for a bounding box $b_t^l$ not associated with any existing object estimators. An object estimator will be terminated if it is not associated with any bounding boxes for $K$ consecutive frames.

\section{Implementation}\label{sec:impl}
Our implementation of \SUM uses R-CNN in a manner that divides object detection problem into two stages: a) image proposal generation, and b) proposal classification. R-CNN was chosen because of its suitability for small datasets and relatively high accuracy. The baseline image proposals generation method used by R-CNN, selective search~\cite{uijlings2013selective}, was replaced for Edge Boxes~\cite{zitnick2014edge} due to its computational efficiency and recall~\cite{hosang2014good}. 
The Edge Box method is an edge-based proposal generation method which applies a score function to evaluate the completeness of contours that contain in a certain bounding box. By implementing structured decision tree, the evaluation process for a huge number of candidate boxes can be performed in a second. A Softmax layer was used for the final label output layer rather than a separate SVM.

For particle filtering, we employed the CUDA-OpenGL interoperation to render all particles in a single render buffer on the GPU and can be accessed by the CUDA kernels to compute the weights for particles. The major computation is operated on directly on GPU and there is very few data transfer between GPU and CPU memory. This provides a tractable solution for us to employ more particles to sample hypothesized objects. 

Manipulation actions used TRAC-IK~\cite{beeson2015trac} to generate the joint states of the arm given the pose of the end-effector and MoveIt!\cite{sucan2013moveit} to perform motion planning afterwards. 
Based on methods proposed by ten Pas and Platt~\cite{ten2016localizing}, a custom grasp planning pipeline was developed to evaluate all possible grasp candidates based on the Darboux frame (surface normal and principal curvature axes) of each object vertex.

\begin{figure}
\centering
\begin{floatrow}
\centering
\subfloat[]{
\includegraphics[width=0.45\columnwidth]{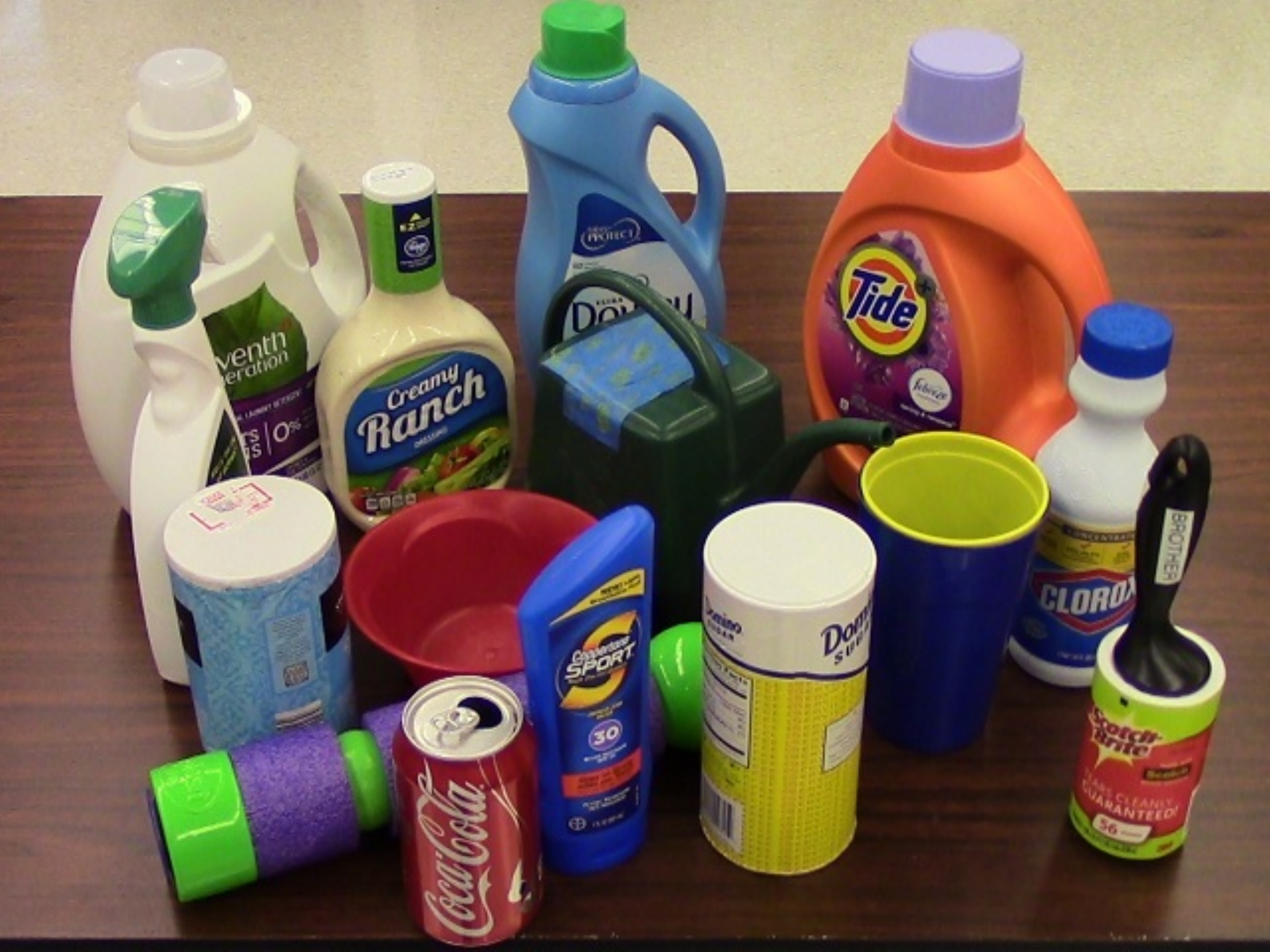}
}

\subfloat[]{
\includegraphics[width=0.45\columnwidth]{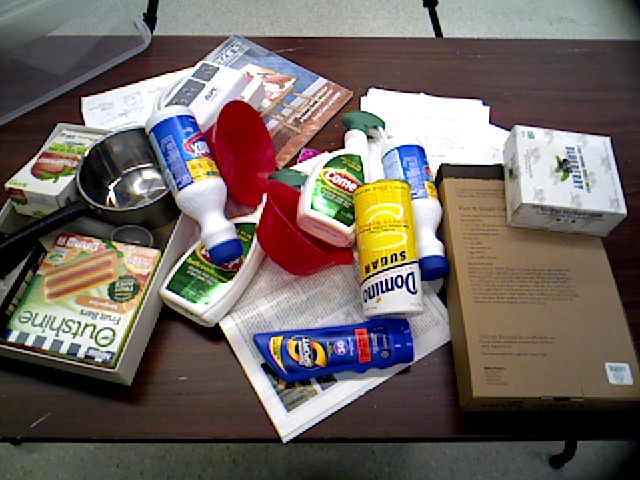}
\label{raw_rgb_image1}
}

\end{floatrow}

\begin{floatrow}
\subfloat[]{
\includegraphics[width=0.45\columnwidth]{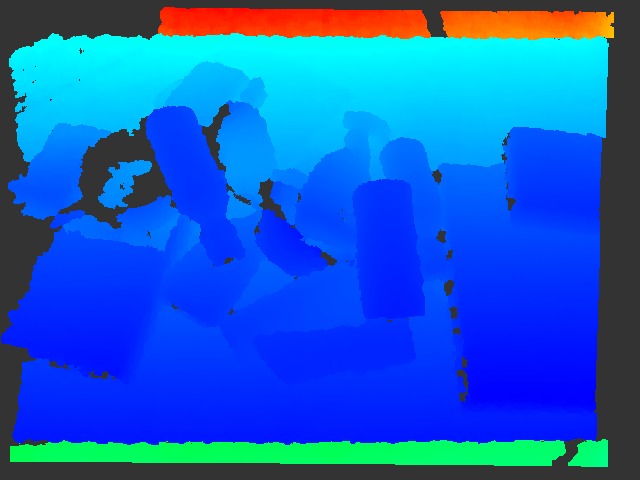}
\label{observed_11}
}
\subfloat[]{
\includegraphics[width=0.45\columnwidth]{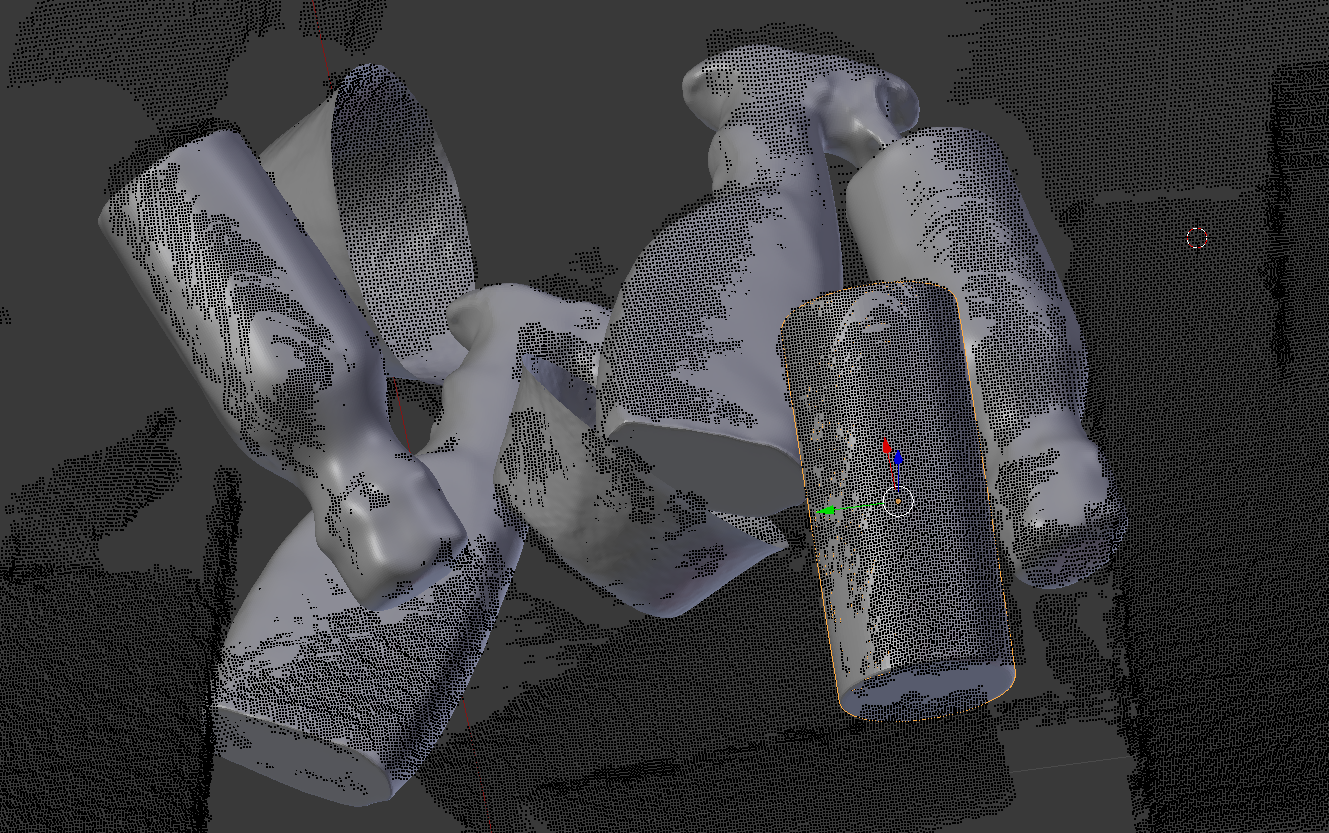}
\label{blender1}
}
\end{floatrow}

\caption{\footnotesize The \SUM dataset objects (a).  Eight of these objects in a cluttered scene (b) viewed as an observed depth image (c) and as ground truth (d). }
\label{fig:progress_dataset}
\end{figure}

\section{Results}

We first examined \SUM on single scenes where estimates static images without robot actions. We compare \SUM with a local descriptor, Fast Point Feature Histograms (FPFH)~\cite{rusu2010fast}, on 10 test scenes of cluttered unstructured environment. For sequential manipulation, eight experiments for sorting objects into two groups were performed with a Fetch mobile manipulation robot. 

\SUM was run on a Ubuntu 14.04 system with an Titan X Graphics card and CUDA 7.5with 625 particles and 25 resampling iterations for all trials. $\sigma_{c}$ is set to 0.1. Sensor resolution $\epsilon$ is set to 0.008 in meters. $\sigma_1$,  $\sigma_2$ and $\sigma_3$ are set to 0.04, 0.02, 0.01 respectively. A custom dataset of 15 household objects (Figure \ref{fig:progress_dataset}) was used for testing, as well as 3D model generation.
For CNN training, 8-10 streams of each object in the dataset was captured in a variety of different poses with different backgrounds. 
The whole training dataset contains 8366 ground truth images and 60563 background images. The Caffenet model~\cite{jia2014caffe} was used for network fine-tuning, which was trained on ImageNet~\cite{imagenet_cvpr09}.

\subsection{Single Scene Estimation}
\begin{figure}
\centering
\begin{floatrow}
\centering
\subfloat[]{
\includegraphics[width=0.47\columnwidth]{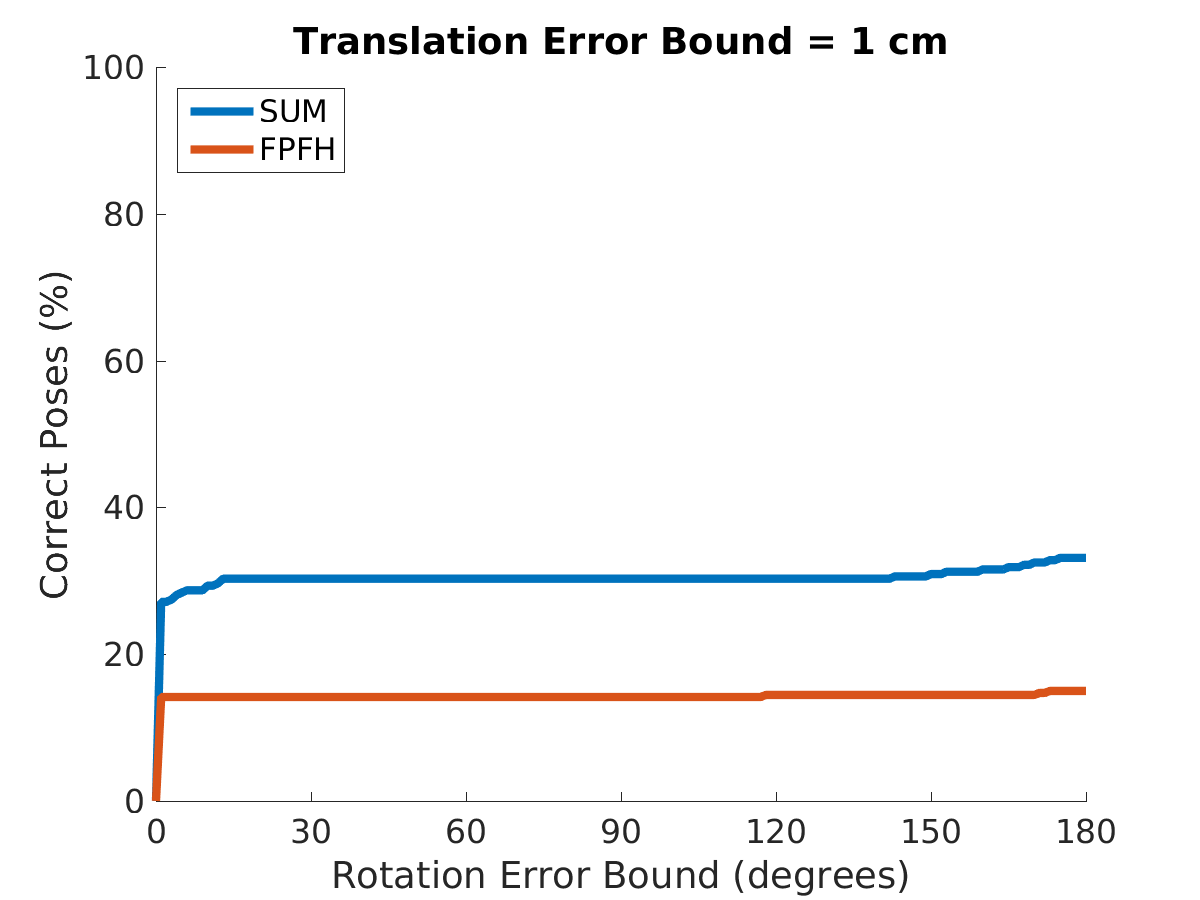}
\label{pose_err_1cm}
}

\subfloat[]{
\includegraphics[width=0.47\columnwidth]{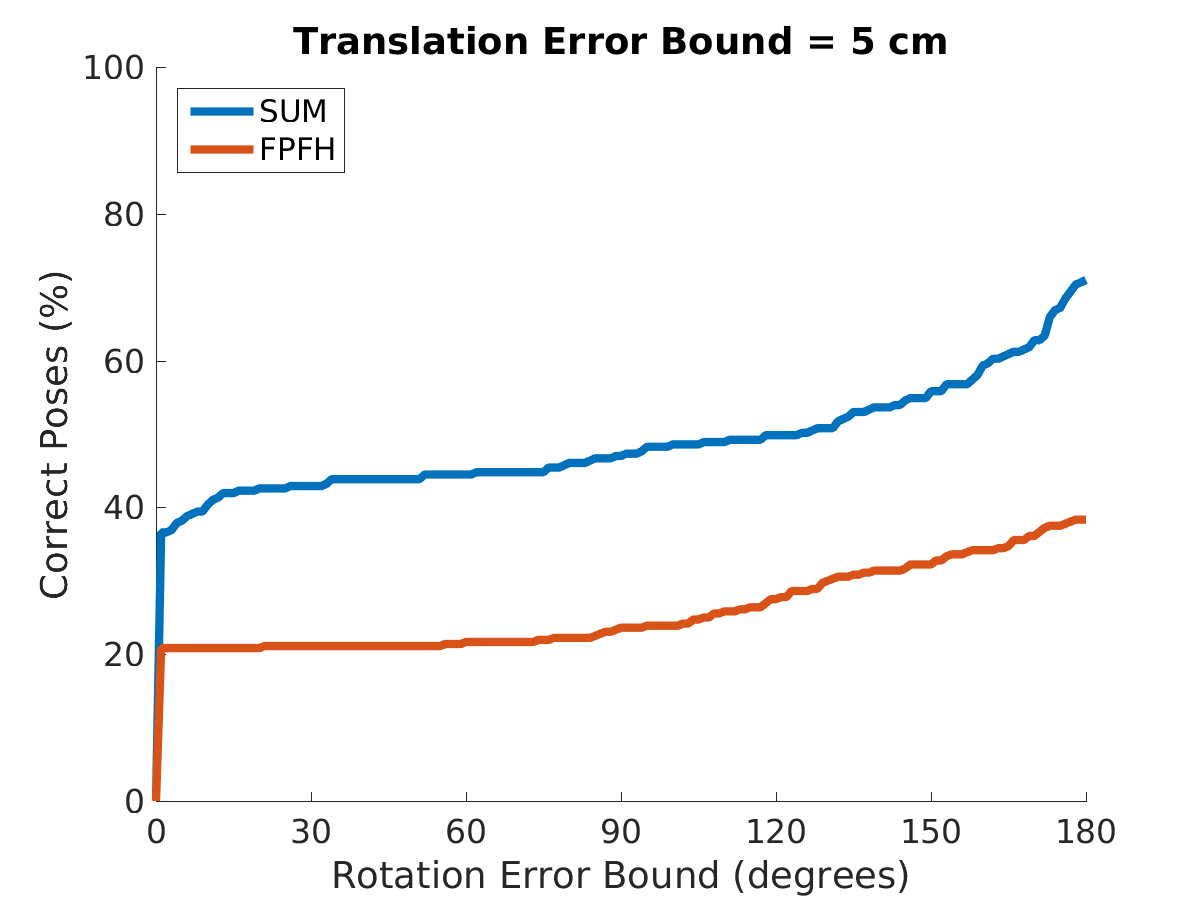}
\label{pose_err_5cm}
}
\end{floatrow}

\begin{floatrow}
\subfloat[]{
\includegraphics[width=0.47\columnwidth]{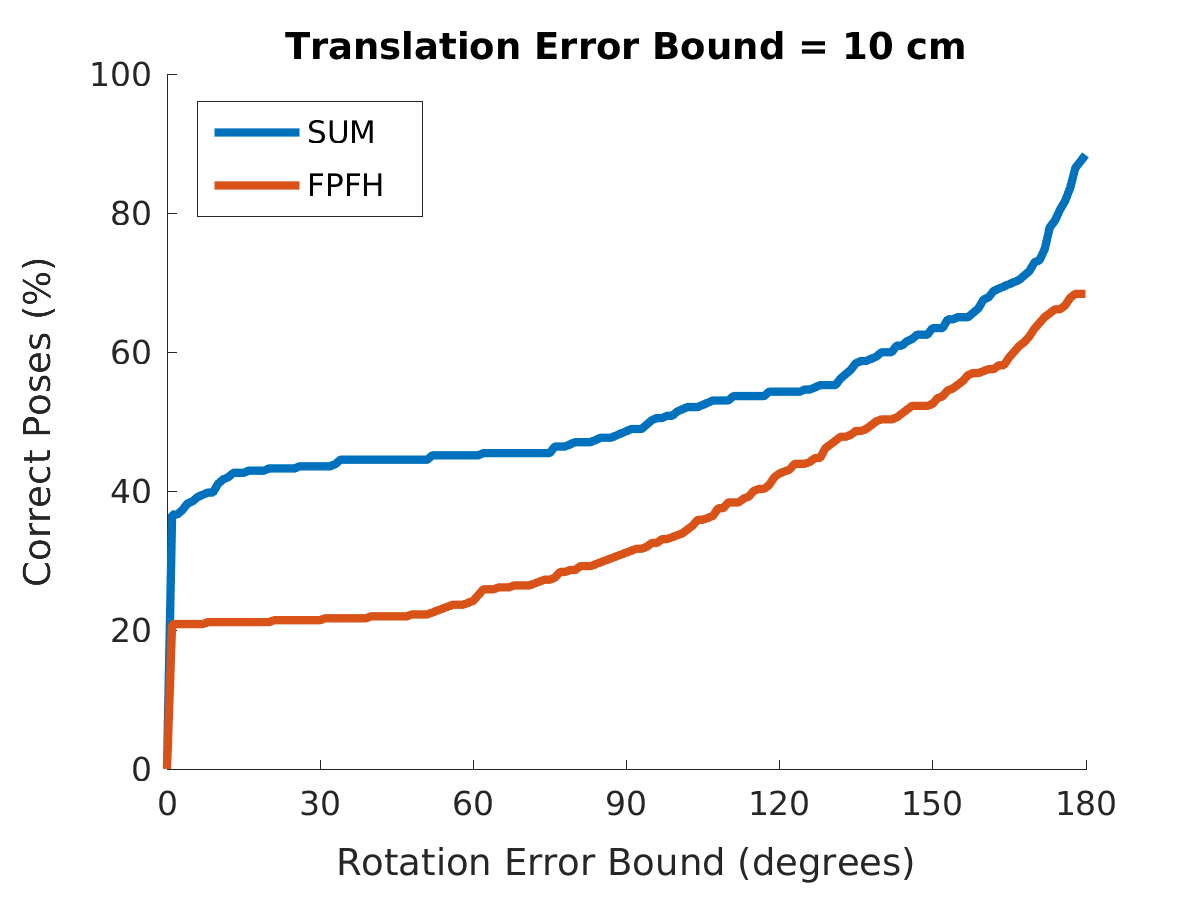}
\label{pose_err_10cm}
}

\subfloat[]{
\includegraphics[width=0.47\columnwidth]{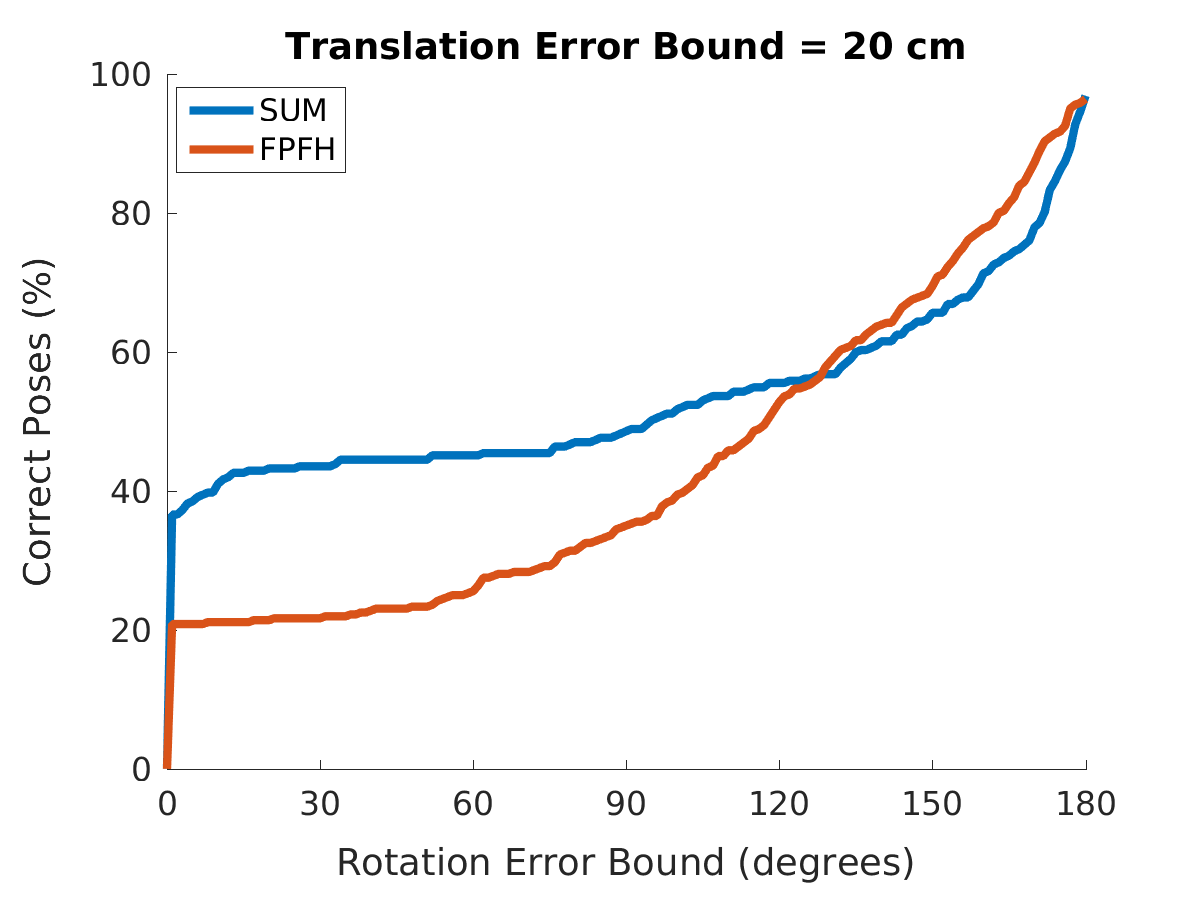}
\label{pose_err_20cm}
}

\end{floatrow}
\caption{\footnotesize The plots compare the performance between our method \SUM and FPFH with respect to the accuracy of correct poses. In each plot, there is a fixed translation error bound (1cm, 5cm, 10cm and 20cm), the x-axis is the changing rotation error bound and the y-axis is the percentage of the correct poses. Each point in the plot shows the accuracy of correctly localized objects with a fixed translation and rotation error bound. }
\label{fig:progress_exp_pose_err}
\end{figure}

\begin{figure*}

\subfloat[\footnotesize]
{
\includegraphics[width=0.3\linewidth]{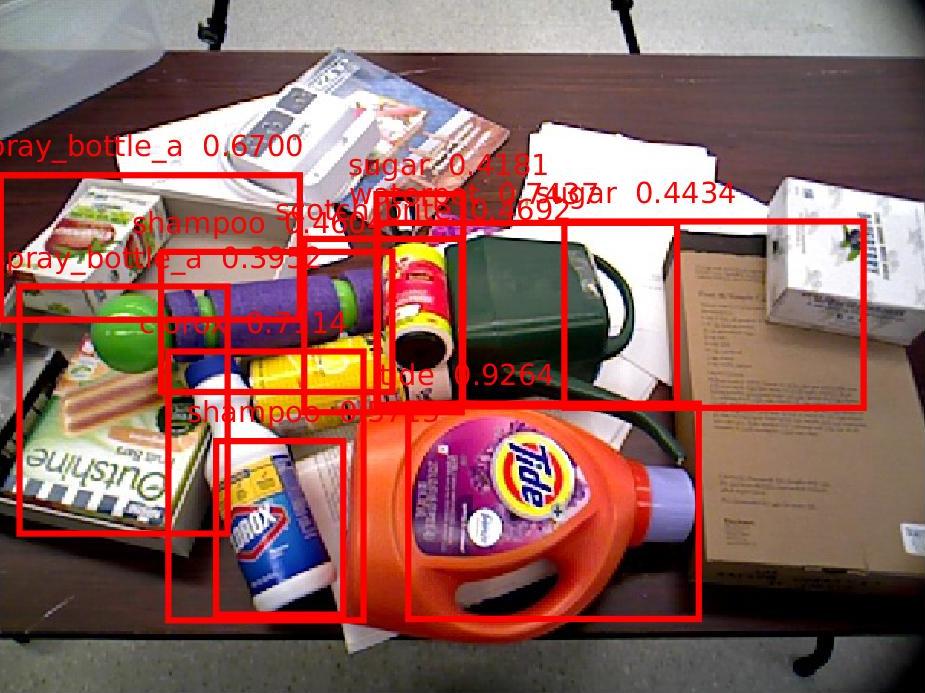}
\label{ex1_detection}
}
\subfloat[\footnotesize]
{
\includegraphics[width=0.3\linewidth]{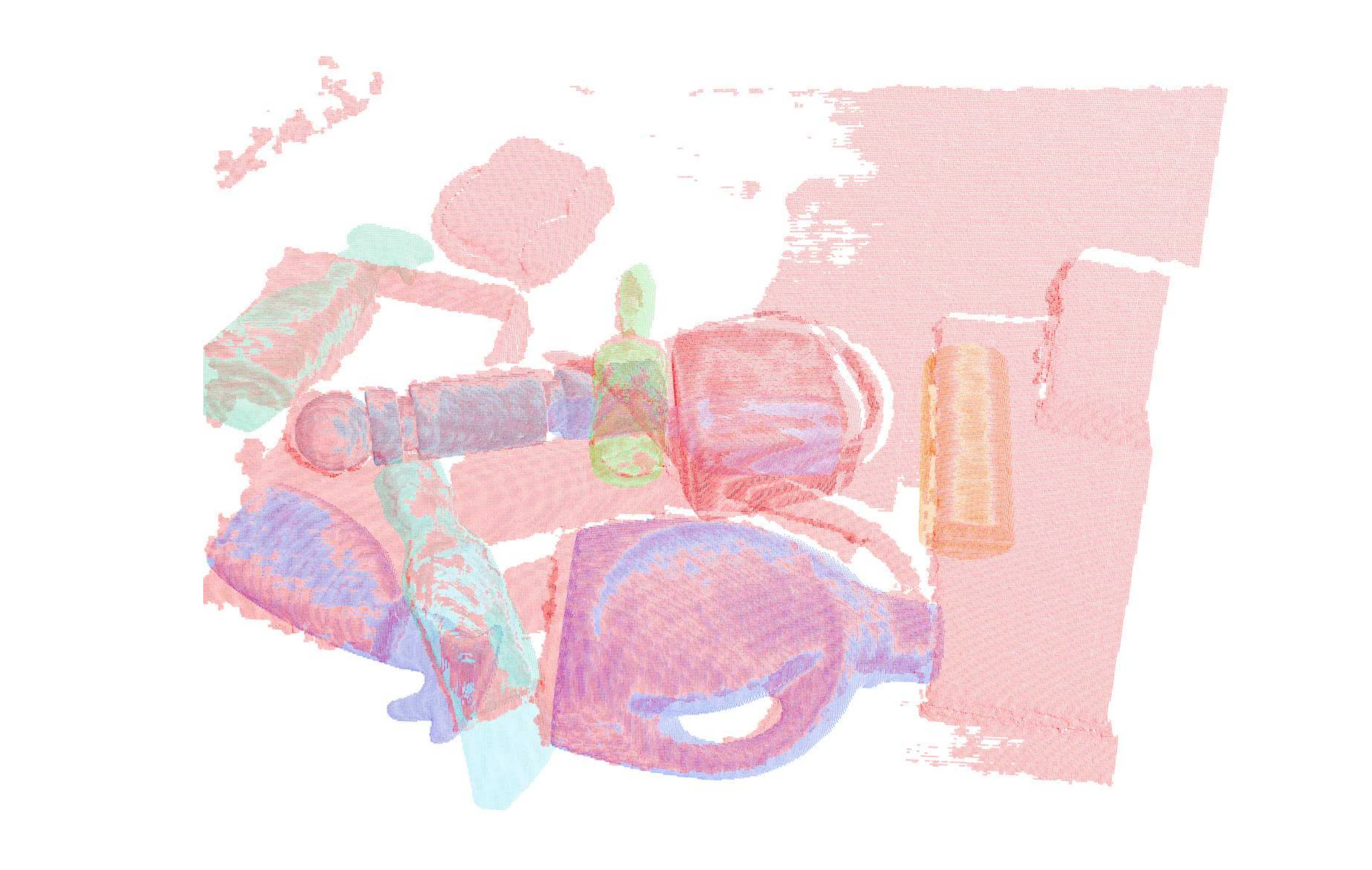}
\label{ex1_all}
}
\subfloat[\footnotesize]
{
\includegraphics[width=0.3\linewidth]{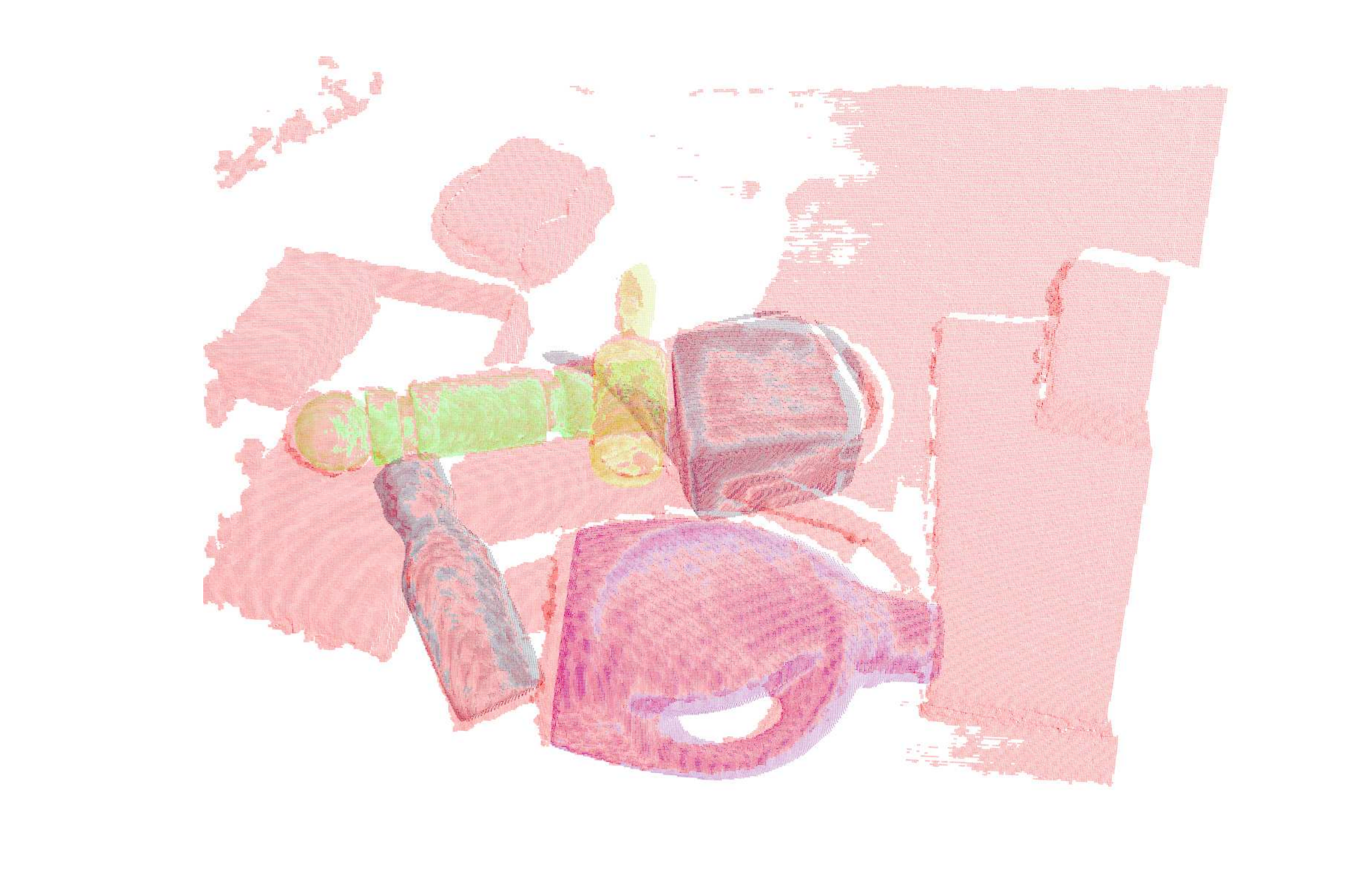}
\label{ex1_corrected}
}

\subfloat[\footnotesize]
{
\includegraphics[width=0.3\linewidth]{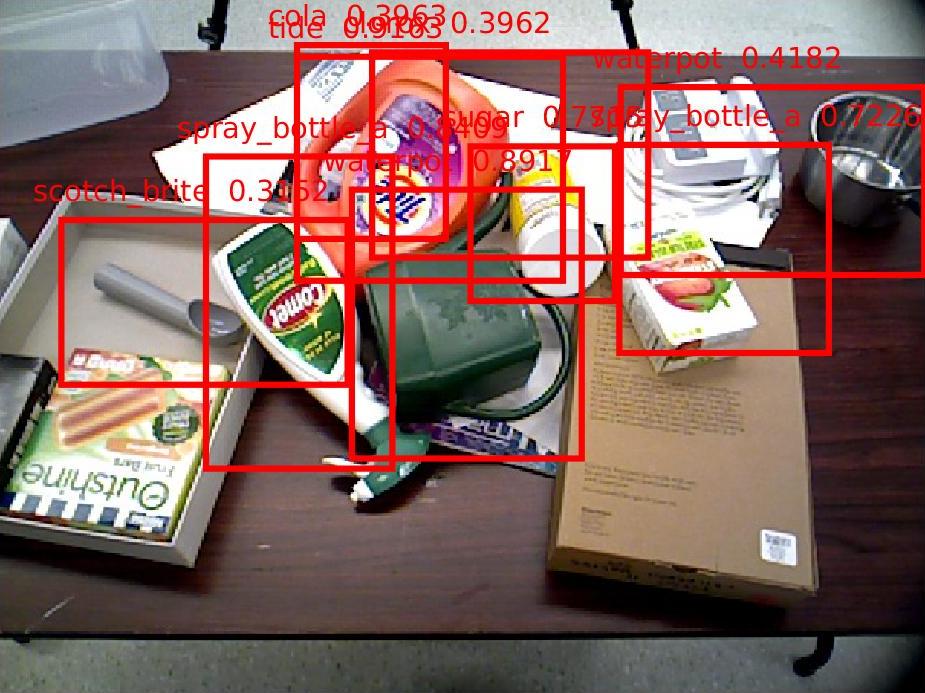}
\label{ex2_detection}
}
\subfloat[\footnotesize]
{
\includegraphics[width=0.3\linewidth]{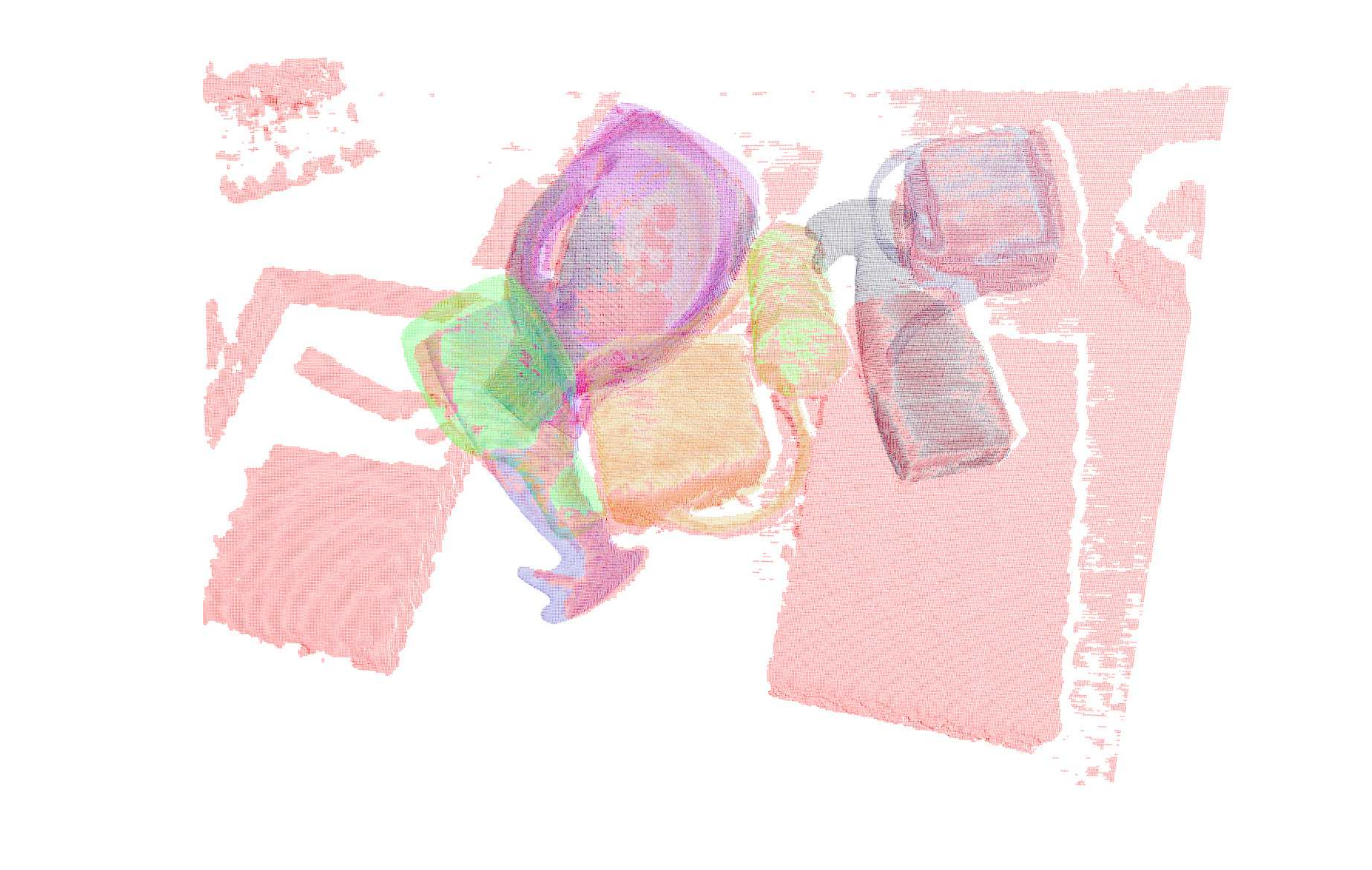}
\label{ex1_all}
\label{ex2_all}
}
\subfloat[\footnotesize]
{
\includegraphics[width=0.3\linewidth]{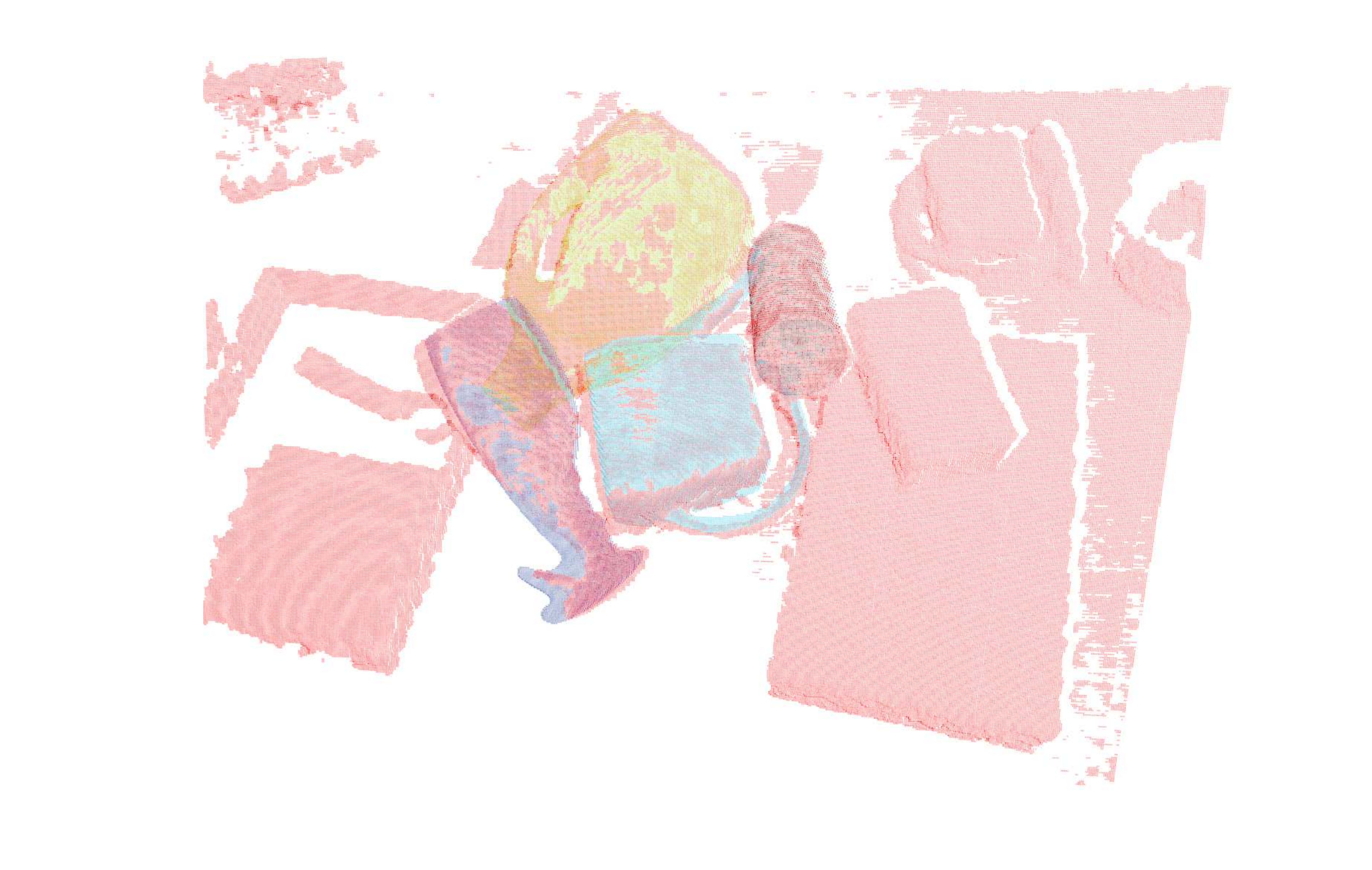}
\label{ex2_corrected}
}
\caption{\footnotesize (a)(d) The detection results from EgdeBox and R-CNN object detector . (b)(e) Intermediate results from \SUM which contain false positive estimation (c) The scene estimate result of (a) after thresholding, correct the false positives: two "spray bottle", a "shampoo" and a "sugar". (f) The scene estimate result of (d), correct the false positives: "ranch", "waterpot", "spray bottle" and "tide".}
\label{fig:correctedfalse}
\vspace{-1em}
\end{figure*}

\SUM was evaluated and compared with FPFH on 10 single scenes, with 10 trials each, with respect to the accuracy of estimated object poses.We compute the accuracy of correct poses over all the test scenes and all the runs. 
Accuracy is defined as the number of correctly localized objects over the total number of detection true positives from the RCNN object detector, where a true positive has IoU greater than 0.5 between estimated and ground truth bounding boxes. 
We deem an object as correctly localized if its translation error and rotation error fall with chosen error bounds. The translation error is the Euclidean distance between estimated object position $(x, y, z)$ and ground truth pose $(x_{gt}, y_{gt}, z_{gt})$ and the orientation error is the shortest angle error between estimate object $(roll, pitch, yaw)$ and ground truth $(roll_{gt}, pitch_{gt}, yaw_{gt})$. 

The four plots in Fig. \ref{fig:progress_exp_pose_err} depict the comparison between \SUM and the baseline method. In each plot, there is a fixed translation error bound (1cm, 5cm, 10cm and 20cm) and the x-axis is the changing rotation error bound. The y-axis is the percentage of the correct poses. We can see in Fig.~\ref{pose_err_1cm}, ~\ref{pose_err_5cm} and~\ref{pose_err_10cm}, our method performs better than FPFH in the small error bounds (translation error smaller than 1cm, 5cm 10cm). 
\SUM can also reject false positives from detection results.  Fig~\ref{fig:correctedfalse} shows two examples of how \SUM corrected false positives from detection results. We also calculated the mean ratio of rejected detection false positives. 
The mean ratio of rejected detection false positives over all the test scenes is 0.84 and the standard error over 10 runs is 0.0126.

\subsection{Estimation and Manipulation on Sequential Scenes}
In the manipulation evaluation, the robot must sort object on a cluttered tabletop into cleaning and non-cleaning categories by picking and placing the object into the right or left bin. In order to make a natural unstructured scene, we avoid manually placing objects in the scene by indiscriminately pouring the objects onto the cluttered table. 
After scene estimation by \SUM, the object with the most likely estimate is selected to be grasped and sorted into the appropriate bin. 
No matter whether the robot succeeds or not, \SUM updates the pose hypotheses by the action model, associates the object estimators with current detection results and estimates the scene iteratively.

The manipulation results are shown in Table~\ref{table:sequential_scenes}. Each scene contains five recognizable objects. We evaluate the method by the completion ratio of each sequence shown in the last row of the table. The completion ratio is how much the recognizable objects on the table are successfully sorted by the robot. The robot successfully completed six out of eight sequences. In sequence(a), failure occured when robot was trying to pick up "downy", it swept "sugar can" onto the ground. In sequence(f), the robot failed to pick up ``waterpot'' as the feasible grasp poses are out of joint limits of the arm.  
As shown in the second row, a manipulation action error occurs about once on average per trial.
Despite such errors, \SUM performs robustly to not only detection uncertainties but also manipulation failures. As shown in Figure~\ref{fig:sequential_scenes}, there is a manipulation error in the fourth action of the sequence, where the ``spray bottle'' slipped from the gripper. \SUM subsequently estimated this object again and the robot picked it up successfully.

\begin{table*}
\centering
\resizebox{\columnwidth}{!}{
\begin{tabular}{c@{\qquad}cccccccc}
  \toprule
  & Sequence(a) & Sequence(b) & Sequence(c) & Sequence(d) & Sequence(e) & Sequence(f) & Sequence(g) & Sequence(h)  \\
  \midrule
  \thead{ Number of \\ total objects} 
  		& 5 & 5 & 5 & 5 & 5 & 5 & 5 & 5 \\[0.1cm]
  \thead{Number of \\ Manipulation Errors} 
  		& 1 & 1 & 2 & 0 & 0 & 1 & 1 & 0 \\[0.1cm]
  \thead{Number of \\ Manipulation Trials} 
  		& 4 & 6 & 7 & 5 & 5 & 5 & 6 & 5 \\[0.1cm]
  \thead{Completion \\ Ratio} 
  		& 0.80 & 1.0 & 1.0 & 1.0 & 1.0 & 0.8 & 1.0 & 1.0 \\[0.1cm]
  \bottomrule
\end{tabular}
}
\caption{\footnotesize The tables shows results of manipulation experiments for 8 sequences. The first row shows the number of object in each scene. Row two and three show the the count for the manipulation errors and trials for the sequence. The last row shows the ratio of how much the recognizable objects on the table are successfully sorted by the robot. }
\label{table:sequential_scenes}
\end{table*}

\newcommand{\subfigwidth}{0.32}
\begin{figure}
\subfloat
{
\includegraphics[width=\subfigwidth\linewidth]{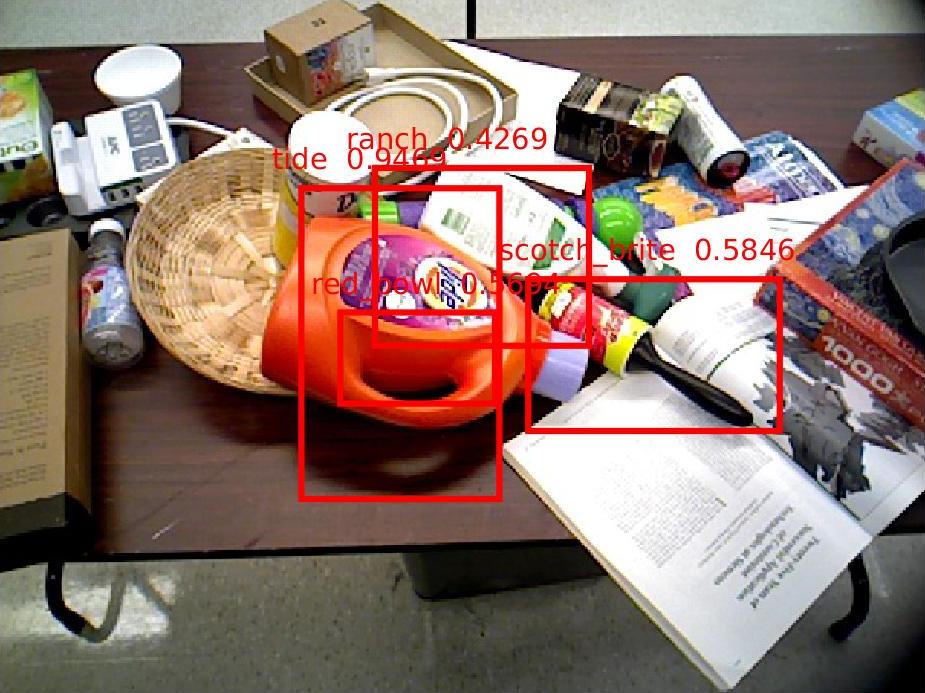}
\label{exp10_1a}
}
\subfloat
{
\includegraphics[width=\subfigwidth\linewidth]{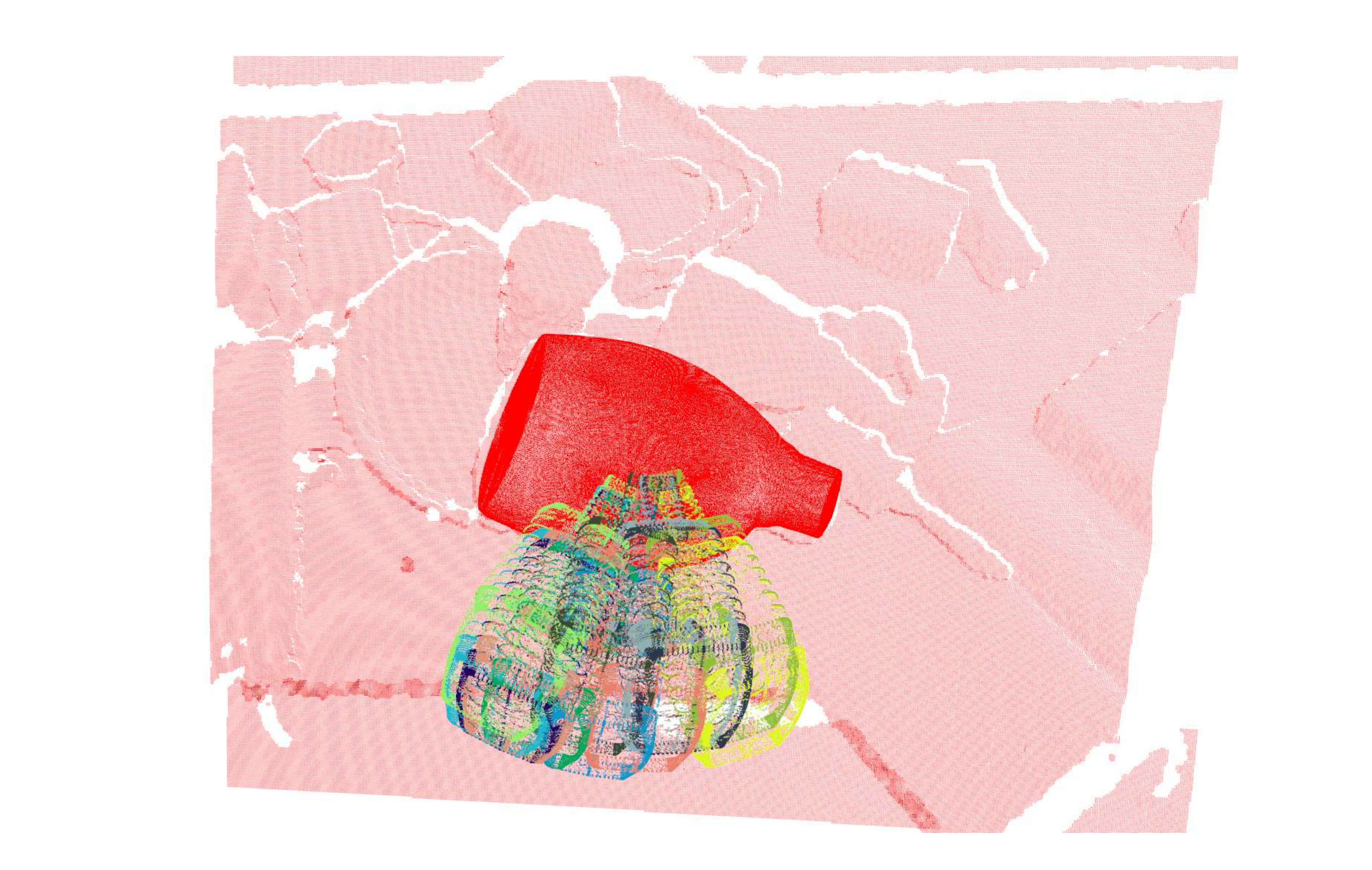}
\label{exp10_1c}
}
\subfloat
{
\includegraphics[width=\subfigwidth\linewidth]{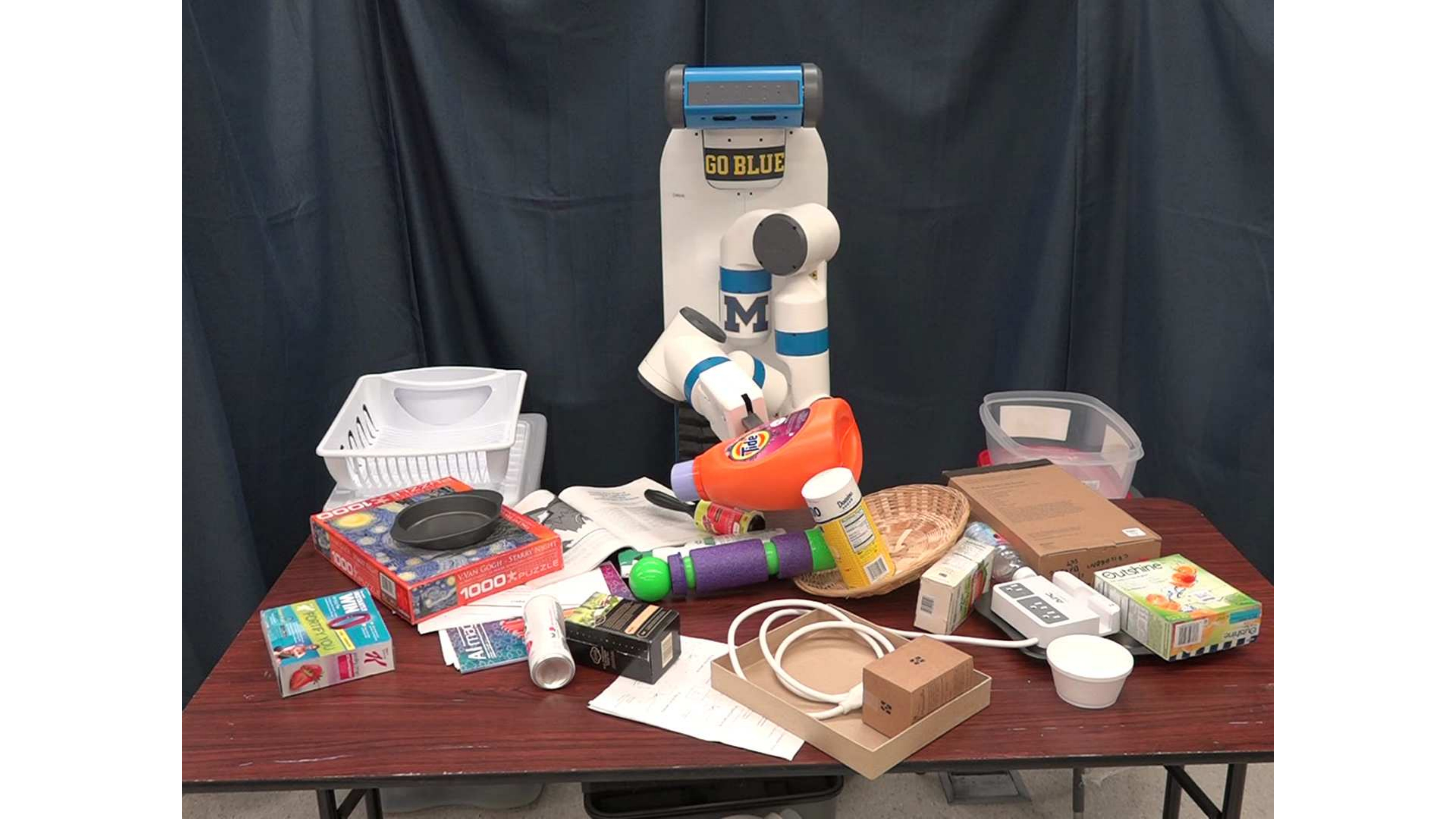}
\label{exp10_1d}
}

\subfloat
{
\includegraphics[width=\subfigwidth\linewidth]{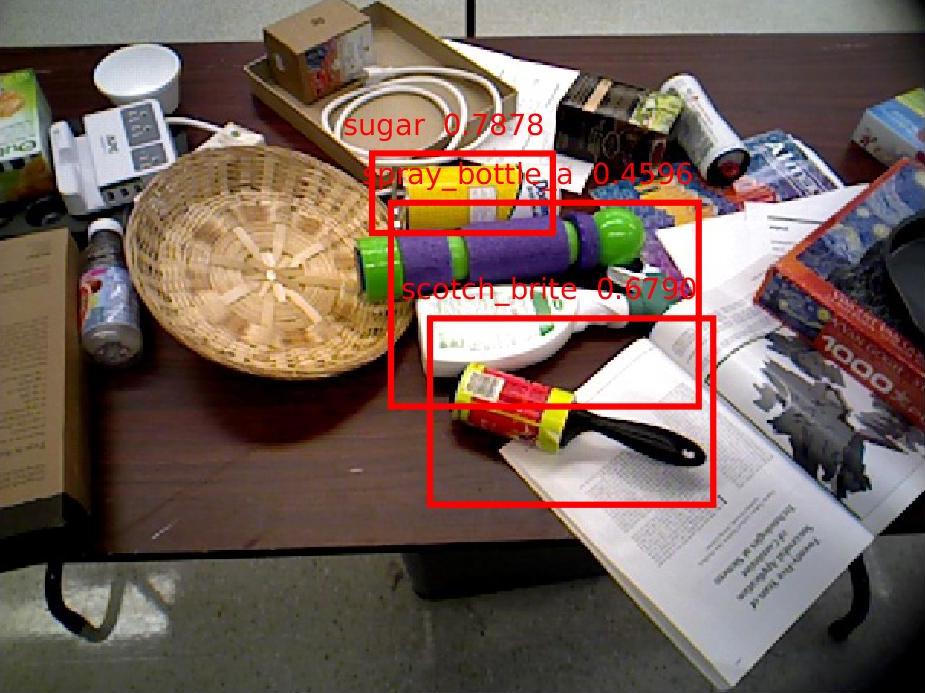}
\label{exp10_2a}
}
\subfloat
{
\includegraphics[width=\subfigwidth\linewidth]{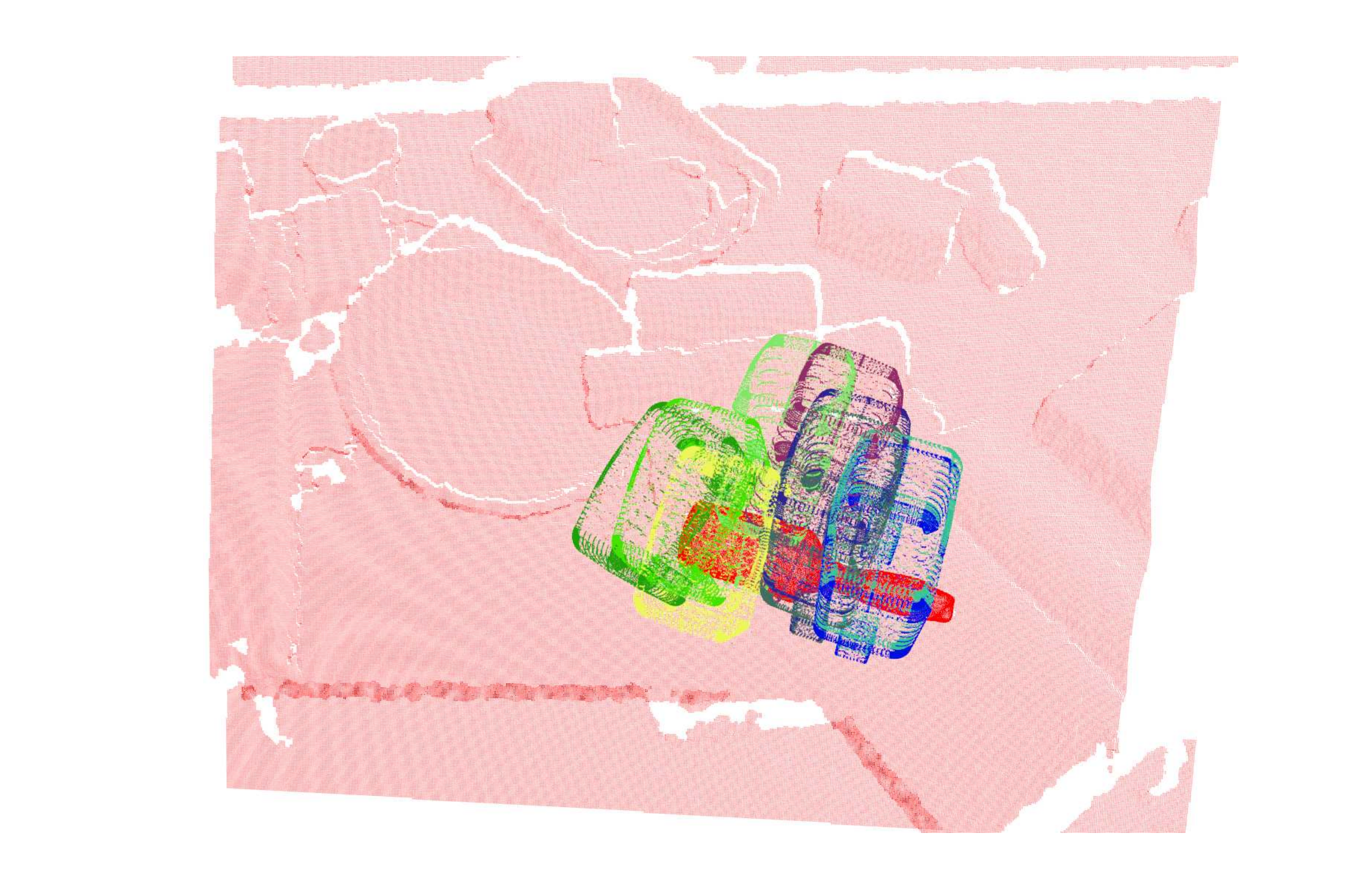}
\label{exp10_2c}
}
\subfloat
{
\includegraphics[width=\subfigwidth\linewidth]{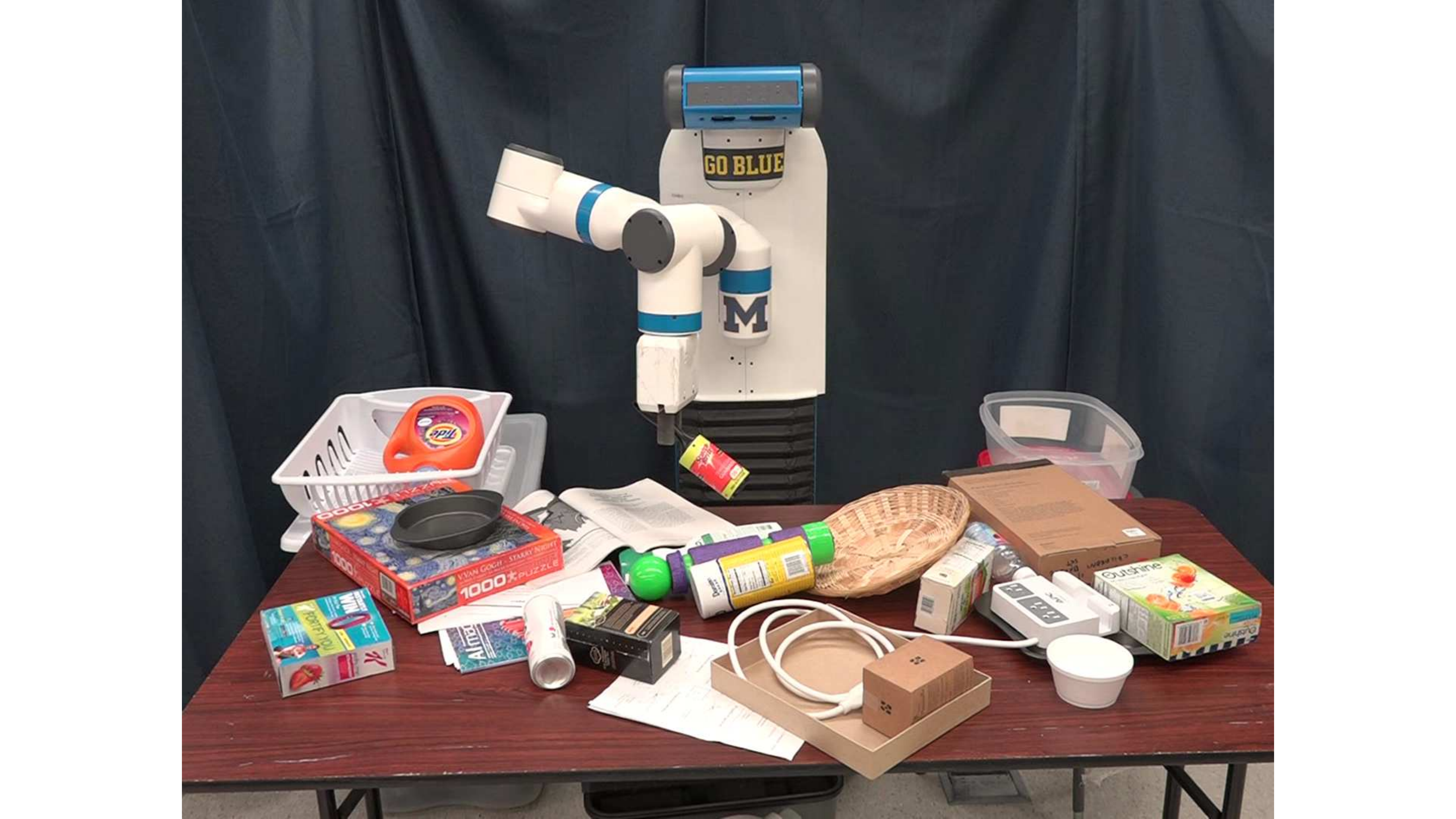}
\label{exp10_2d}
}

\subfloat
{
\includegraphics[width=\subfigwidth\linewidth]{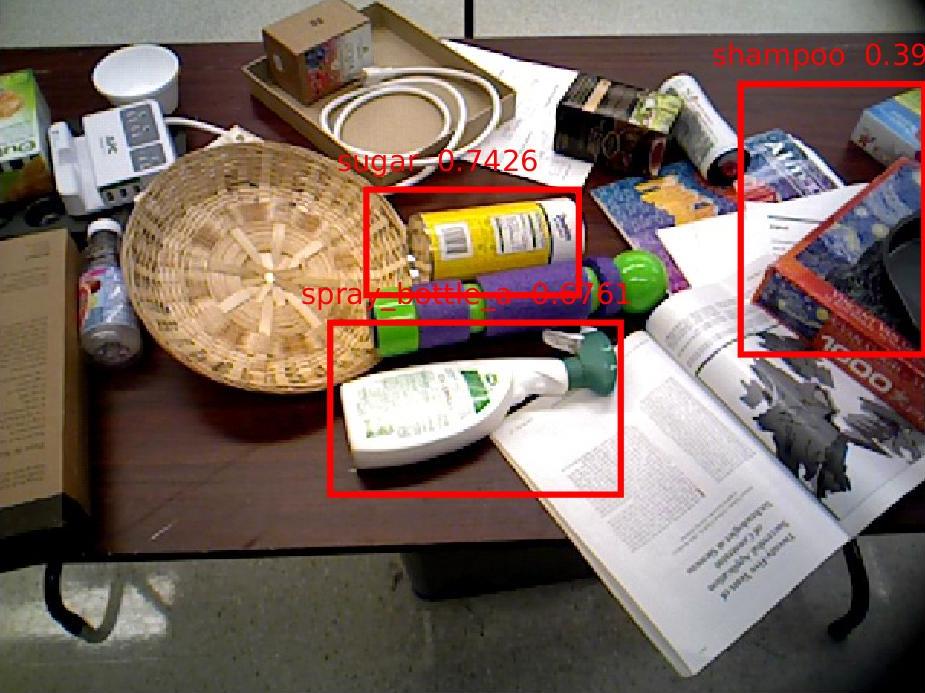}
\label{exp10_3a}
}
\subfloat
{
\includegraphics[width=\subfigwidth\linewidth]{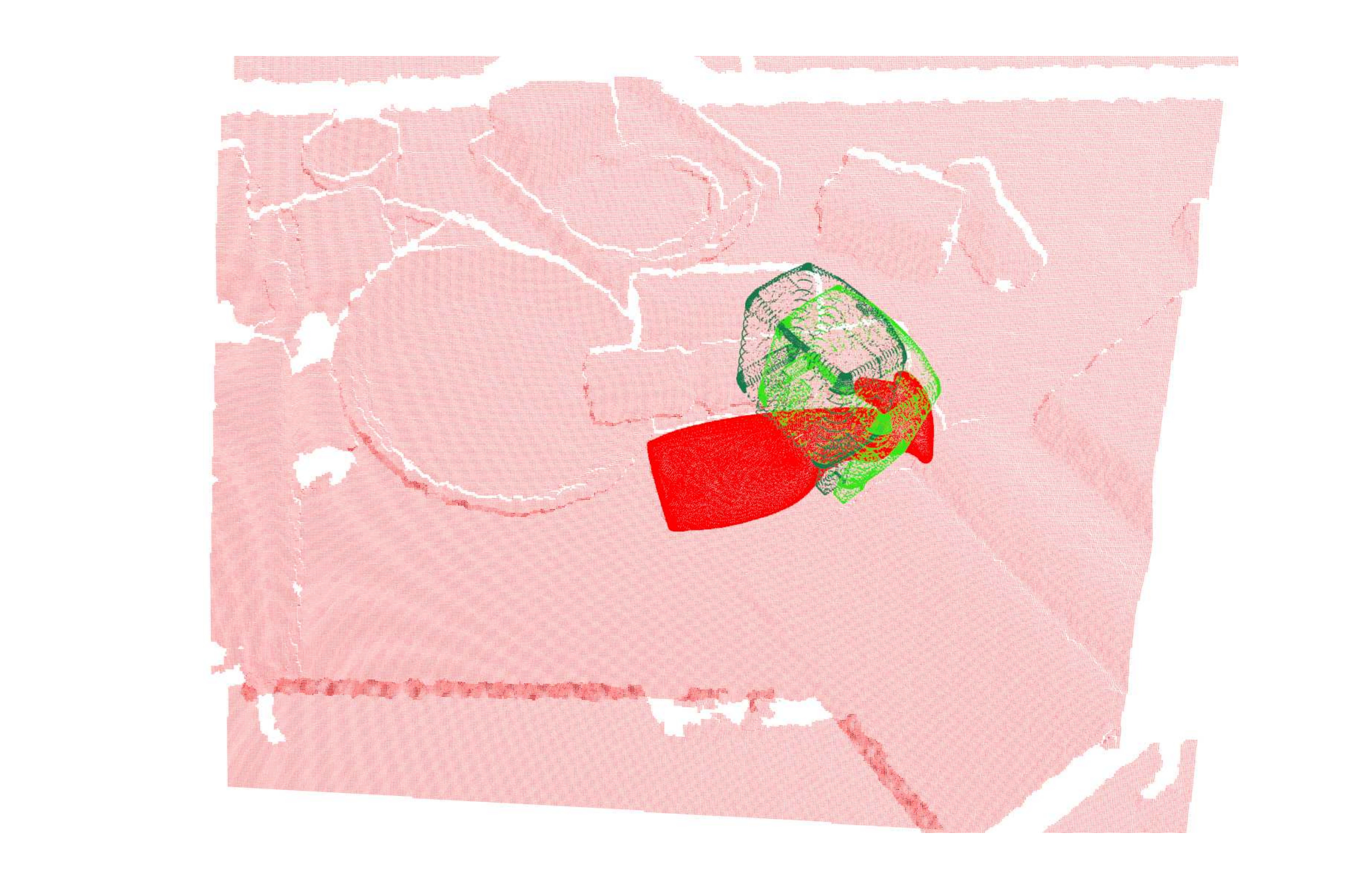}
\label{exp10_3c}
}
\subfloat
{
\includegraphics[width=\subfigwidth\linewidth]{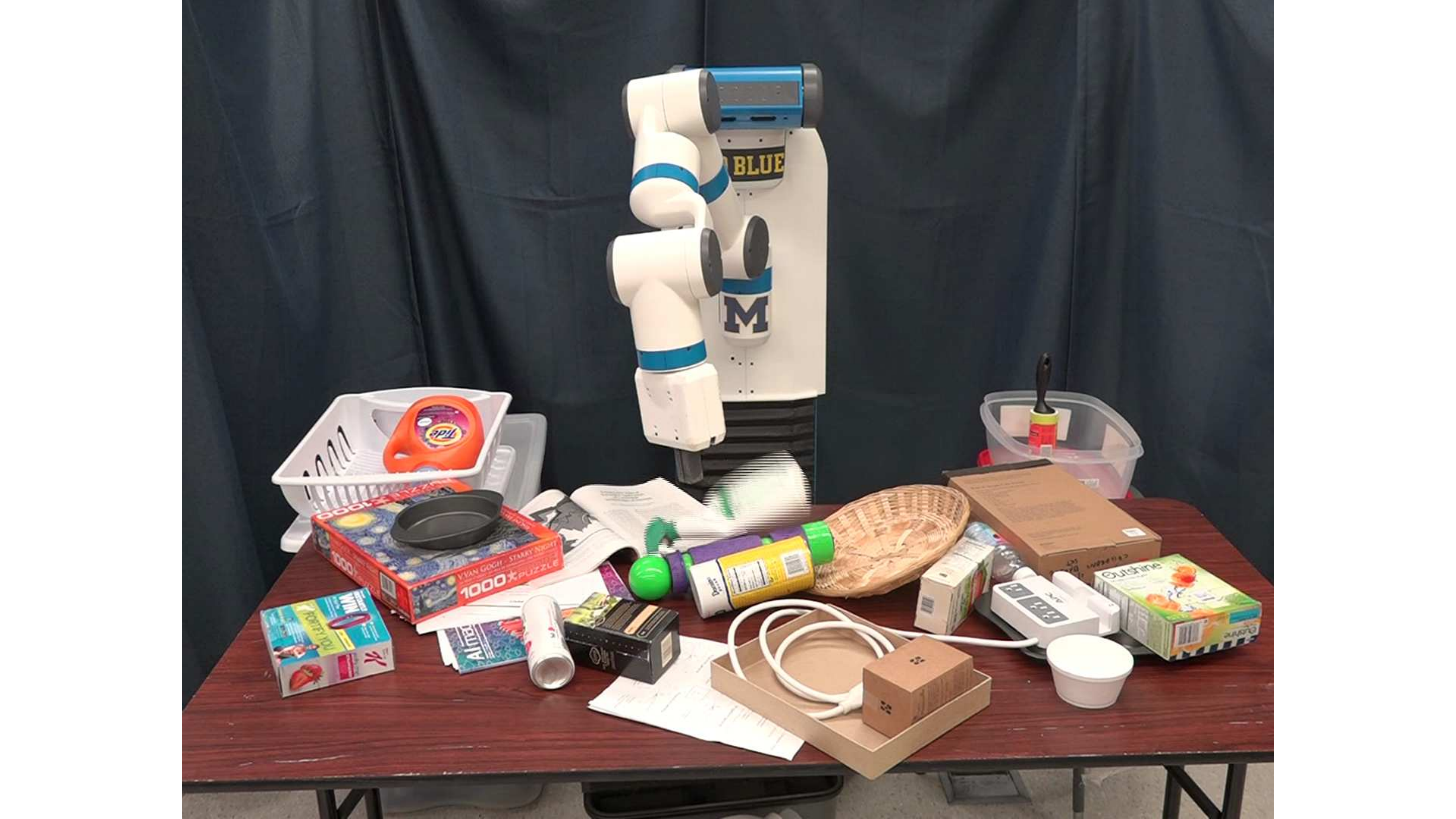}
\label{exp10_3d}
}

\subfloat
{
\includegraphics[width=\subfigwidth\linewidth]{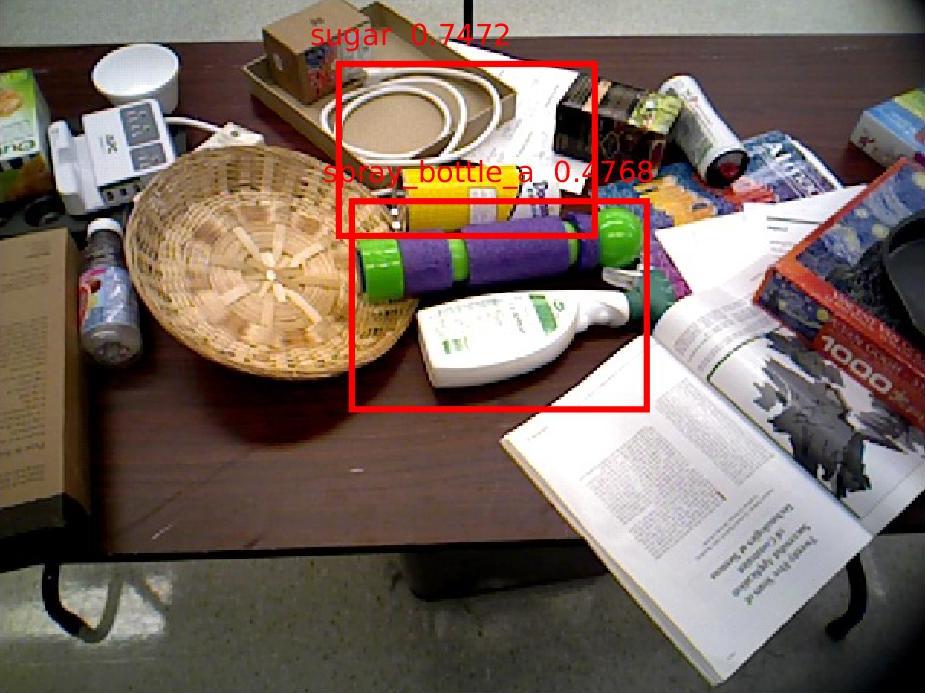}
\label{exp10_4a}
}
\subfloat
{
\includegraphics[width=\subfigwidth\linewidth]{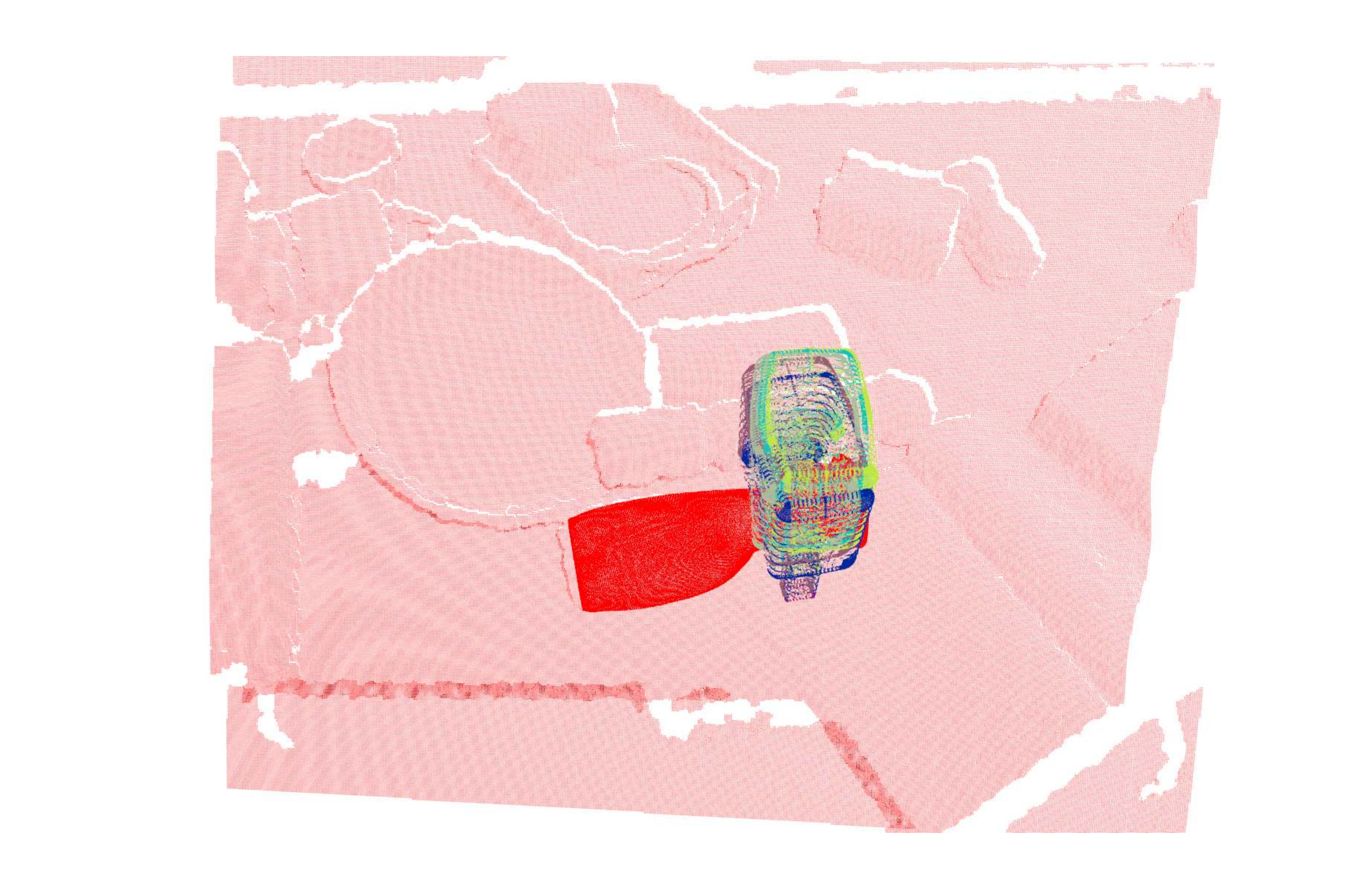}
\label{exp10_4c}
}
\subfloat
{
\includegraphics[width=\subfigwidth\linewidth]{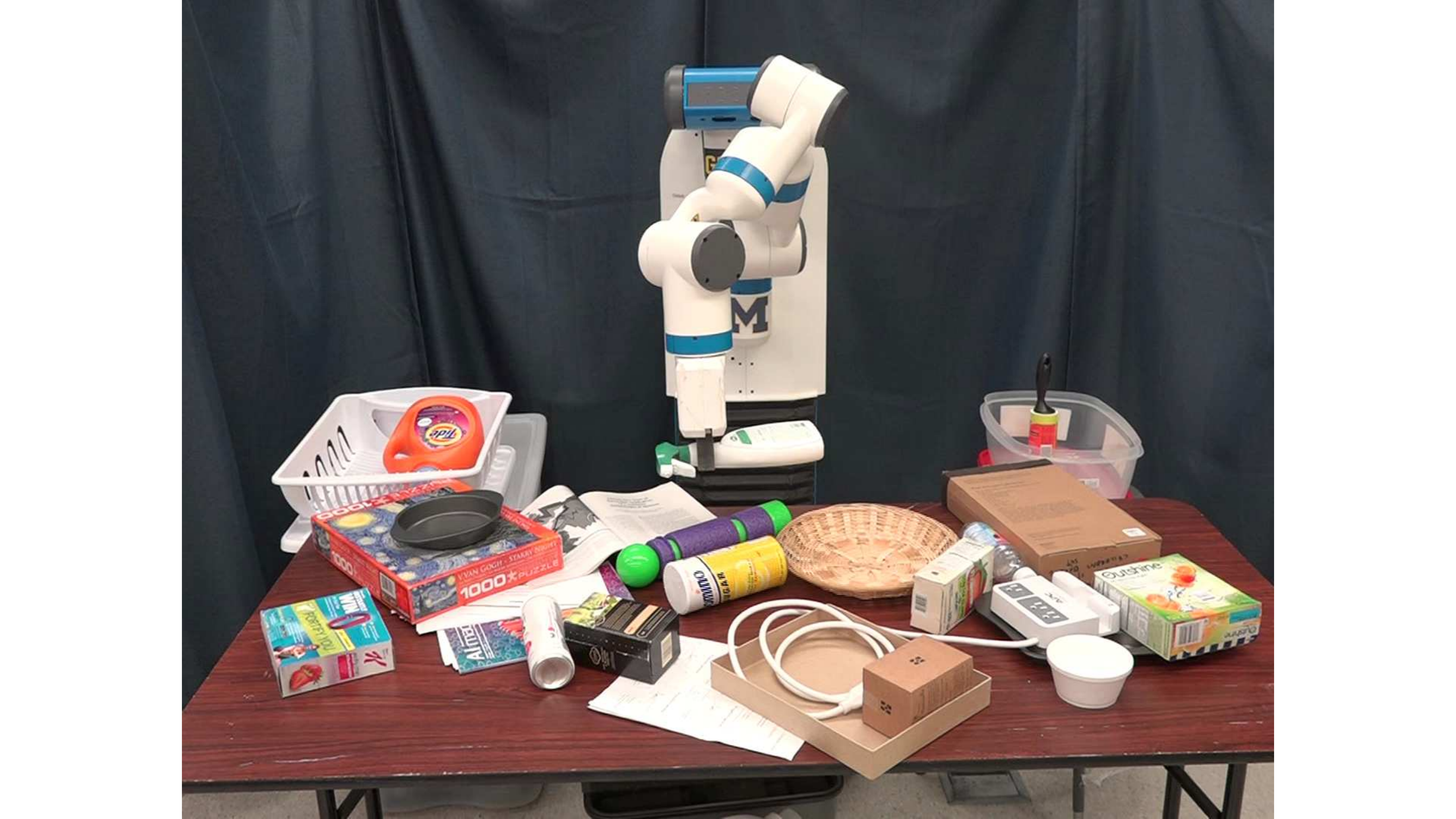}
\label{exp10_4d}
}

\subfloat
{
\includegraphics[width=\subfigwidth\linewidth]{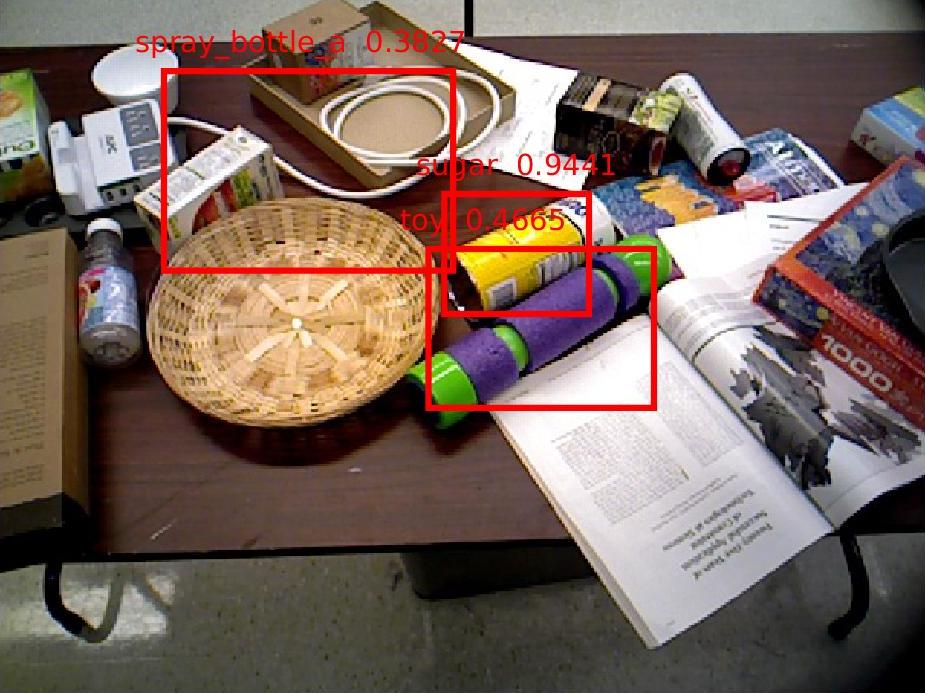}
\label{exp10_5a}
}
\subfloat
{
\includegraphics[width=\subfigwidth\linewidth]{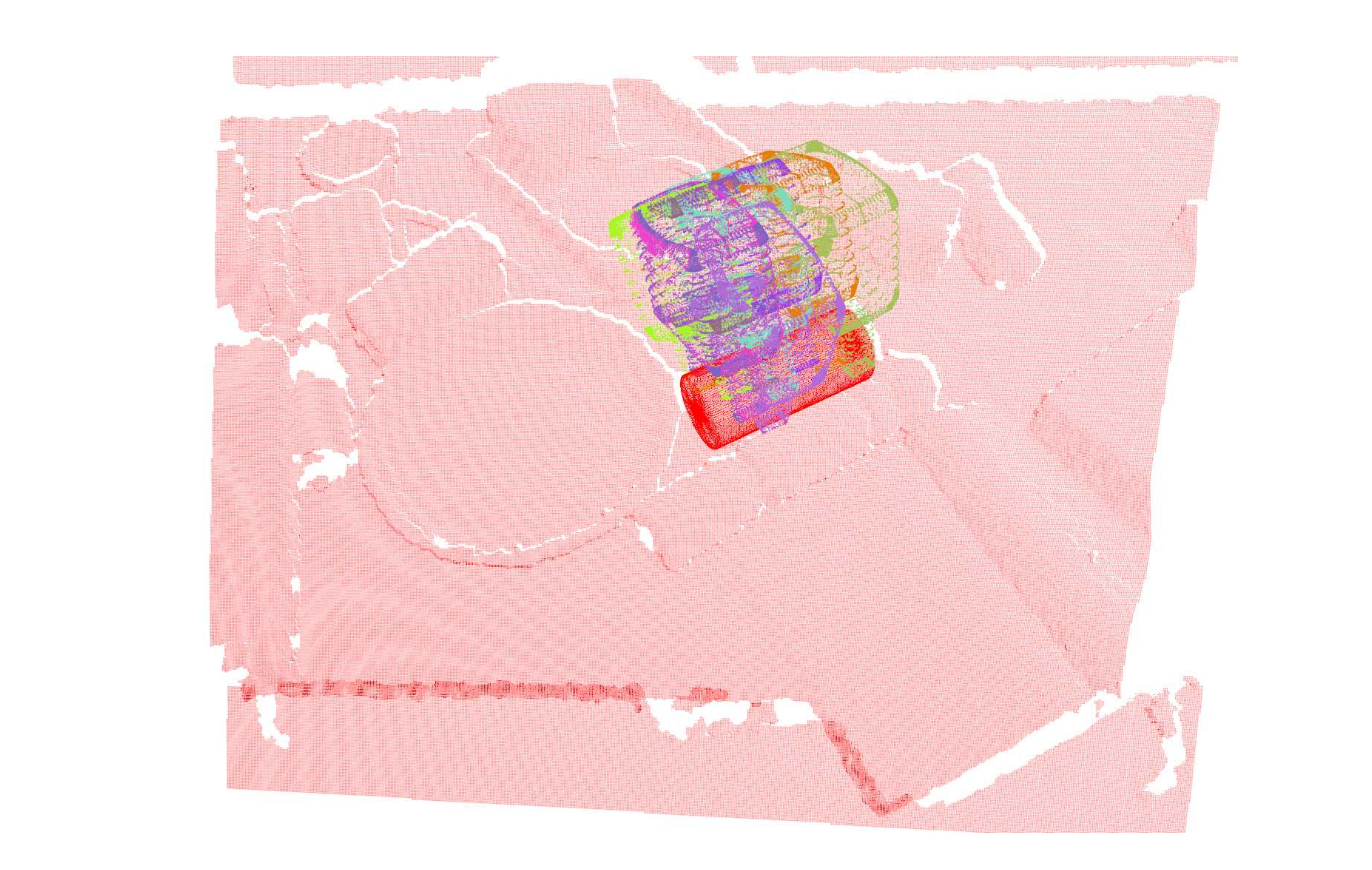}
\label{exp10_5c}
}
\subfloat
{
\includegraphics[width=\subfigwidth\linewidth]{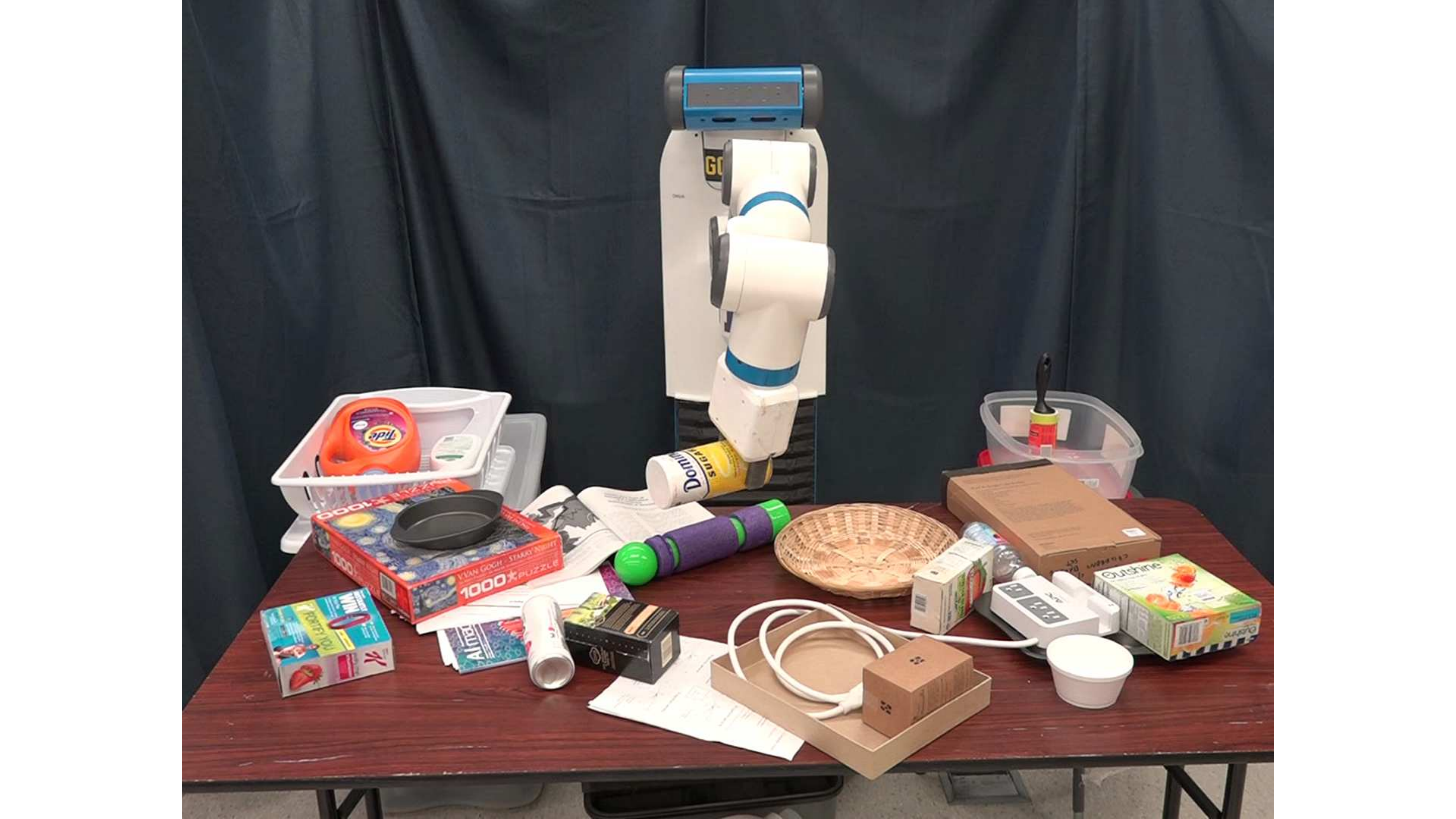}
\label{exp10_5d}
}

\subfloat
{
\includegraphics[width=\subfigwidth\linewidth]{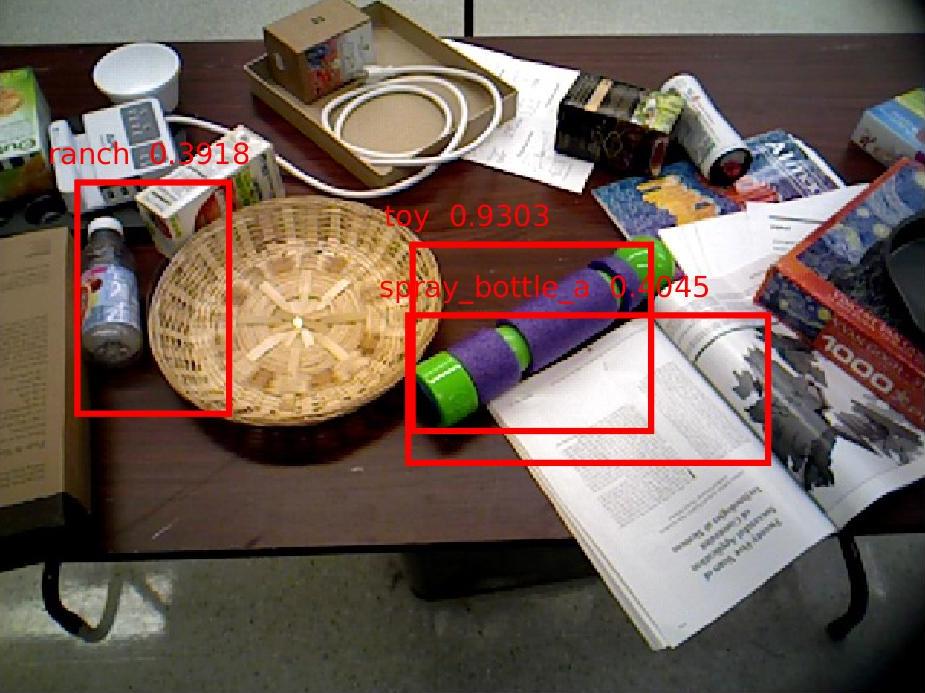}
\label{exp10_6a}
}
\subfloat
{
\includegraphics[width=\subfigwidth\linewidth]{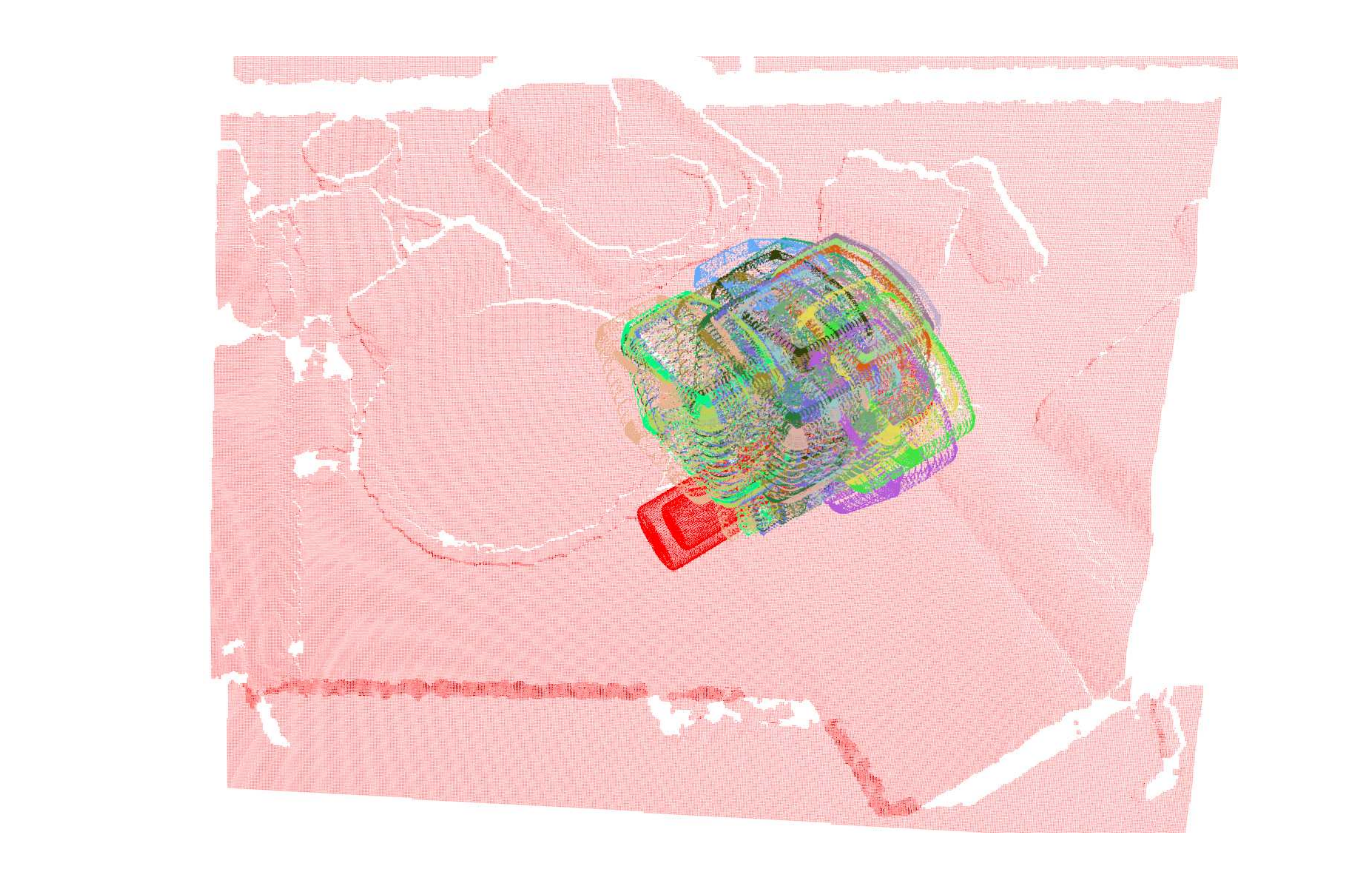}
\label{exp10_6c}
}
\subfloat
{
\includegraphics[width=\subfigwidth\linewidth]{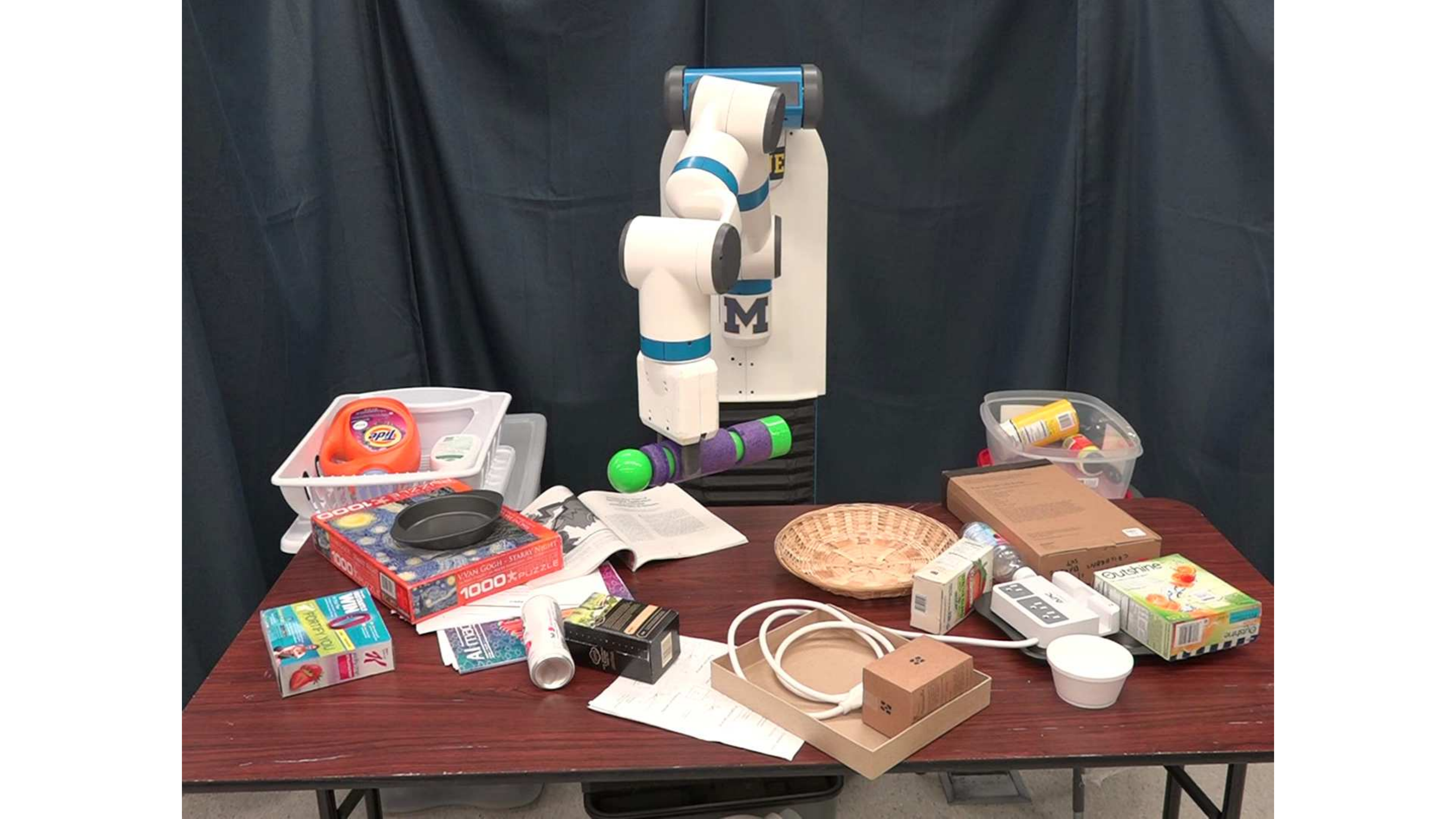}
\label{exp10_6d}
}

\caption{\footnotesize Sequential manipulation by a robot to sort 5 objects on a cluttered tabletop into two bins: cleaning (right bin) and non-cleaning (left bin). 
From the left to right is the detection results from RCNN object detector, most likely object from \SUM, computed collision-free grasp poses and robot manipulating the object in action. Our system estimated and manipulated ``tide'', ``scotch brite'', ``spray bottle'', ``sugar'' and ``toy'' sequentially. 
}
\label{fig:sequential_scenes}
\end{figure}

\section{Conclusion}
In this paper, we propose \SUM as a combined generative and discriminative approach to robust sequential scene estimation and manipulation. \SUM utilizes output from discriminative object detector and recognizer to guide the generative process of sampling scene hypothesis for 6DOF pose estimation. By maintaining a belief over object poses over a sequence robot actions, \SUM is able to perform robust estimation and manipulation in a cluttered and unstructured tabletop scenario.







\bibliographystyle{abbrv}
\bibliography{big}

\end{document}